\documentclass[letterpaper]{article}
\usepackage{uai2019}
\usepackage[margin=1in]{geometry}

\usepackage{caption}
\usepackage{subcaption}

\usepackage{amsmath,amscd,amssymb}
\usepackage{amsfonts}
\usepackage[pdftex]{graphicx}
\usepackage{array}
\usepackage{bbold}
\usepackage{bm}
\usepackage[table,dvipsnames]{xcolor}
\usepackage{tabu}
\usepackage{multirow}
\usepackage{makecell}

\usepackage{mathtools}

\usepackage[round]{natbib}
\DeclarePairedDelimiter{\ceil}{\lceil}{\rceil}

\usepackage[shortlabels]{enumitem}
\setlist[enumerate]{itemsep=0.25pt}

\usepackage{mathtools}

\usepackage{times}

\usepackage[colorlinks]{hyperref}
\hypersetup{citecolor=SkyBlue,linkcolor=Periwinkle}

\title{Neural Likelihoods for Multi-Output Gaussian Processes}

\author{} 

\author{ {\bf Martin Jankowiak\thanks{\;\;Correspondence to: jankowiak@uber.com}} \\
Uber AI Labs \\
San Francisco, CA, USA
\And
{\bf Jacob Gardner}  \\
Uber AI Labs \\
San Francisco, CA, USA}

\newcommand{\NN}{\mathcal{N}}

\newcommand{\bff}{\mathbf{f}}
\newcommand{\bfft}{\tilde{\mathbf{f}}}
\newcommand{\bmm}{\mathbf{m}}
\newcommand{\bu}{\mathbf{u}}

\newcommand{\by}{\mathbf{y}}

\newcommand{\bx}{\mathbf{x}}
\newcommand{\bz}{\mathbf{z}}

\newcommand{\bX}{\mathbf{X}}
\newcommand{\bZ}{\mathbf{Z}}

\newcommand{\bsig}{\bm{\sigma}}
\newcommand{\bSig}{\bm{\Sigma}}

\newcommand{\beps}{\bm{\epsilon}}
\newcommand{\bPhi}{\mathbf{\Phi}}

\newcommand{\bmu}{\bm{\mu}}

\newcommand{\bF}{\mathbf{F}}
\newcommand{\bY}{\mathbf{Y}}

\newcommand{\bM}{\mathbf{M}}
\newcommand{\bMt}{\widetilde{\mathbf{M}}}
\newcommand{\bFt}{\widetilde{\mathbf{F}}}
\newcommand{\bW}{\mathbf{W}}
\newcommand{\bbeta}{\bm{\beta}}

\newcommand{\erf}{{\rm erf}}
\newcommand{\lgc}{\cellcolor[gray]{0.95}}
\newcommand{\RR}{\mathbb{R}}
\newcommand{\EE}{\mathbb{E}}

\newcommand{\mean}{\mathsf{m}}
\newcommand{\var}{\mathsf{v}}
\newcommand{\bmean}{\boldsymbol{\mathsf{m}}}
\newcommand{\bvar}{\boldsymbol{\mathsf{v}}}

\newif\ifcomments
\commentstrue
\ifcomments\newcommand{\comments}[1]{#1}\else\newcommand{\comments}[1]{}\fi

\begin{document}

\maketitle

\begin{abstract}
We construct flexible likelihoods for multi-output Gaussian process models that leverage neural networks as components.
We make use of sparse variational inference methods to enable scalable approximate inference for the resulting class of models.
An attractive feature of these models is that they can admit analytic predictive means even when the likelihood is non-linear and the predictive distributions are non-Gaussian. We validate the modeling potential of these models in a variety of experiments in both the supervised and unsupervised setting. We demonstrate that the flexibility of these `neural' likelihoods can improve prediction quality as compared to simpler Gaussian process models and that neural likelihoods can be readily combined with a variety of underlying Gaussian process models, including deep Gaussian processes.
\end{abstract}

\section{Introduction}

Significant effort has gone into developing flexible, tractable probabilistic models, especially for the supervised settings of
regression and classification.
These include, among others, multi-task Gaussian processes \citep{bonilla2008multi}, Gaussian process regression networks \citep{wilson2011gaussian}, deep Gaussian processes \citep{damianou2013deep},
Gaussian processes with deep kernels \citep{wilson2016deep, calandra2016manifold}, as
well as various approaches to Bayesian neural networks \citep{graves2011practical,blundell2015weight,hernandez2015probabilistic}.
While neural networks promise considerable flexibility, scalable learning algorithms for Bayesian neural networks that can deliver robust uncertainty estimates remain elusive.

For this reason Gaussian processes (GPs) are an important class of models in cases where predictive uncertainty
estimates are important. Gaussian processes offer several key advantages over other probabilistic models:\footnote{We
refer the reader to \citep{rasmussen2003gaussian} for a general introduction to GPs.}
i) the covariance functions they employ typically have a semantic meaning that is natural for practitioners to reason about;
ii) they facilitate incorporating prior knowledge;
and iii) they tend to yield high-quality uncertainty estimates, even for out-of-sample data. These strengths mirror corresponding weaknesses of current approaches to Bayesian neural networks, weaknesses that become especially evident in the small data regime, where Bayesian neural networks often struggle to deliver meaningful uncertainties. Despite these strengths, the simplest variants of GP models often fall short of the flexibility of their neural network counterparts.

In recent years, a number of researchers have formulated more flexible Gaussian process models by modifying the GP prior itself.
One approach has been to define richer classes of kernels. This approach is exemplified by deep kernels, which use a deep neural network to define a rich parametric family of kernels \citep{wilson2016deep, calandra2016manifold}. Another complementary approach is the use of deep Gaussian processes, which compose multiple layers of latent functions to build up more flexible---in particular non-Gaussian---function priors \citep{damianou2013deep}. Surprisingly little attention, however, has been paid to the flexibility of \emph{likelihoods} in this setting.
This is likely because, historically, the likelihoods used in the multi-output setting have often been constrained for computational reasons.
However, with recent advances in stochastic gradient methods, some of these structural assumptions are no longer required to enable
efficient inference.

With this opportunity in mind, our aim
 in this work  is to make multi-output Gaussian process models more flexible by equipping them with more flexible likelihoods.
We employ two simple modeling patterns to construct richer likelihoods. To make these modeling patterns more concrete, let $\bF(\bx)$ denote the latent vector of function values drawn from a multi-output GP prior evaluated at an input $\bx$. In the first approach, we pass $\bF(\bx)$ through what is in effect a single-layer neural network before adding Gaussian observation noise. Alternatively, in a second approach we multiply $\bF(\bx)$ by a matrix of coefficients controlled by a deterministic neural network that depends on the inputs to the GP. This latter approach---which can be viewed as a semi-Bayesian version of the Gaussian Process Regression Network \citep{wilson2011gaussian}---is more flexible but is potentially more prone to overfitting.

We find that both modeling patterns result in flexible models that admit efficient inference using sparse variational methods.
Furthermore, we demonstrate empirically that this added flexibility can lead to considerable gains in predictive performance.
Importantly, these \emph{neural} likelihoods are complementary to other approaches for making GP models more flexible, including, for example, deep Gaussian processes and deep kernels.

The rest of this paper is organized as follows.
In Sec.~\ref{sec:related} we place our work in the context of related work.
In Sec.~\ref{sec:models} we describe the models with neural likelihoods that are the focus of this work.
In Sec.~\ref{sec:inf} we describe scalable inference algorithms for this class of models.
In Sec.~\ref{sec:exp} we demonstrate the modeling potential of neural likelihoods with a series of experiments.

\section{Related Work}
\label{sec:related}

As discussed in the introduction, a large body of work aims to make GP
priors more flexible, including deep Gaussian processes \citep{damianou2013deep},
GPs with deep kernels \citep{wilson2016deep, calandra2016manifold},
recurrent Gaussian processes \citep{mattos2015recurrent},
spectral mixture kernels \citep{wilson2013gaussian},
and compositional kernels \citep{sun2018differentiable}.
In the same spirit, a variety of GP models have been proposed
that model correlations between multiple outputs \citep{alvarez2009sparse,alvarez2011computationally}.
In particular these include a number of models that have been formulated in the multi-task setting \citep{bonilla2008multi,williams2009multi,nguyen2014collaborative},
including models for Bayesian optimization \citep{swersky2013multi}.
As mentioned in the introduction, much of this work makes particular structural assumptions
about the covariance structure and/or likelihood for computational convenience; this limits the flexibility of these models.
In this context see \citep{dezfouli2015scalable} for an application of sparse variational methods to a broader class of
likelihoods. 
Finally, \citet{snelson2004warped} construct flexible likelihoods in the GP setting by 
warping the observed outputs with a learned deterministic bijection.\footnote{We experimented with similar constructions but found them to perform poorly in the 
multi-output setting, suffering from a tendency to get stuck in bad local optima.} A similar model, where the warping function is modeled by a GP, is considered in \citep{lazaro2012bayesian}. Our N-MOGP model is closest to this latter setup, with the difference that we work in the multi-output setting and our warping function is provided by a Bayesian neural network.

A number of researchers have explored models that combine various aspects of GPs and neural networks.
For example,  \citet{cutajar2017random} use random feature expansions to formulate a link
between deep GPs and Bayesian neural networks that then enables efficient inference in the resulting class of models.
In \citep{ma2018variational}, the authors propose implicit stochastic processes as a framework for defining flexible function priors; similarly,
Neural Processes are a recent class of models that combine aspects of stochastic processes with neural networks \citep{garnelo2018conditional,garnelo2018neural}. Both these classes of models generally do not employ explicit kernel functions as is characteristic of GPs.
Finally, another work with close analogs to deep kernels is \citep{huang2015scalable}.


\section{Models}
\label{sec:models}

In this section we define the class of models that is the focus of this work.
First, in Sec.~\ref{sec:regmodels} we equip Gaussian process regression models with neural likelihoods.
Next, in Sec.~\ref{sec:lvmmodels} we repurpose a subset of the same models for the unsupervised setting.

Throughout we use  the following notation.
In the regression setting we suppose we are given a dataset $\mathcal{D} = \{ (\bx_i, \by_i) \}_{i=1}^N$ of size $N$ with each input
$\bx_i \in \RR^{D_X}$ and each output $\by_i \in \RR^{D_Y}$. We use $\bX$ and $\bY$ to
refer to the full set of inputs and outputs, respectively.
In the unsupervised setting we assume a dataset $\mathcal{D} = \{  \by_i \}_{i=1}^N$.

\subsection{Models for Regression}
\label{sec:regmodels}

In Sec.~\ref{sec:mogp}-\ref{sec:dgp} we specify three baseline GP models. Then in Sec.~\ref{sec:sbgprn}-\ref{sec:ndgp} we  modify and/or extend these baseline models to obtain four models with neural likelihoods that will form the basis of our experiments.

\subsubsection{Multi-Output Gaussian Processes}
\label{sec:mogp}
We begin by defining our simplest baseline model, a basic multi-output Gaussian process ({\bf MOGP}).\footnote{Compare to the model
in \citep{seeger2005semiparametric}.} We
define $L$ independent Gaussian processes $\{ \bff_\ell(\bx)\}$ with $\ell = 1, ..., L$ and each
with kernel $K_\ell$.\footnote{In general we assume that each $K_\ell$ has its own kernel hyperparameters;
we specify when this is not the case.} For a given input $\bx$ we use $\bF(\bx)$ to denote the $L$-dimensional latent vector of GP function values at $\bx$. The marginal probability of the MOGP is then specified as follows\
\vspace{-1mm}
\begin{equation}
\label{eqn:mogp}
\begin{split}
p(\bY | \bX) = &\!\int \! d\bM p(\bM) \prod_{\ell=1}^L d\bff_\ell p(\bff_\ell | \bX) \times \\
&\prod_{i=1}^N  \NN(\by_i | \bM \bF( \bx_i), \bbeta)
\end{split}
\end{equation}
where $\bM$ is a $D_Y \times L$ mixing matrix, $\bM \bF$ denotes matrix multiplication, and $p(\bM)$ is a unit Normal
prior on $\bM$. Here and throughout $\bbeta$ is a $D_Y$-dimensional vector of precisions that controls the (diagonal) observation noise

Note that for fixed $\bM$ the covariance structure of the $D_Y$-dimensional vector $\bM \bF( \bx)$ is that of the
`linear model of coregionalization' (LMC) \citep{alvarez2012kernels}. While other covariance structures for the Gaussian process prior are possible, for uniformity---and since our primary interest is to investigate modifications to the likelihood---all our models
employ this basic pattern. For the same reason we use RBF kernels throughout.

\subsubsection{Gaussian Process Regression Networks}
\label{sec:gprn}

A natural generalization of the model in Eqn.~\ref{eqn:mogp} is the Gaussian Process Regression Network ({\bf GPRN})
\citep{wilson2011gaussian}. In effect we promote $\bM$ to a $\bx$-dependent matrix of Gaussian processes to obtain
a model
\begin{equation}
\label{eqn:gprn}
\begin{split}
 P(\bY|\bX) =&\int \!  \Pi_{\ell=1}^L d\bff_\ell p(\bff_\ell | \bX)
 \Pi_{m_\ell=1}^{D_Y} d\bmm_{m_\ell}  p(\bmm_{m_\ell} | \bX) \times \\
 &\prod_{i=1}^N  \NN(\by_i | \bM(\bx_i) \bF( \bx_i), \bbeta)
 \end{split}
\end{equation}
where $\bM(\bx_i)$ is now a $D_Y \times L$ random variable governed by a GP prior.\footnote{Following
\citep{wilson2011gaussian} we share kernel hyperparameters among the Gaussian processes $\{\bmm_{m_\ell}\}$
but maintain individual kernels for the $L$ Gaussian processes $\{\bff_\ell\}$. In addition each kernel $K_\ell$ for the latent
function $\bff_\ell$ includes a diagonal noise component.}
Note that this model utilizes $(D_Y + 1) \times L$ Gaussian processes
and so we generally expect inference to be expensive for this class of models.

\subsubsection{Two Layer Deep Gaussian Processes}
\label{sec:dgp}

We consider a deep multi-output GP with two layers of latent functions \citep{damianou2013deep}
\begin{equation}
\label{eqn:dgp}
\begin{split}
 P(\bY|\bX) \!=\!\!\int \!  d\bM p(\bM) &\prod_{\ell=1}^L d\bff_\ell p(\bff_\ell | \bX)
 \!\prod_{\ell^\prime=1}^{L^\prime} d\bfft_{\ell^\prime} p(\bfft_{\ell^\prime} | \bff_{1:L}) \times \\
 &\prod_{i=1}^N  \NN(\by_i | \bM \bFt( \bx_i), \bbeta)
 \end{split}
\end{equation}
where $\bFt(\bx)$ is the $L^\prime$-dimensional vector of Gaussian process function values at $\bx$.
Here $\bM$ is a $D_Y \times L^\prime$ matrix and $p(\bM)$ denotes a unit Normal prior.\footnote{Another alternative
would be to choose $L^\prime = D_Y$ and
set $\bM \to \mathbb{1}$. Since, however, we are particularly interested in the regime where $D_Y$ could be quite high-dimensional---and because inference quickly becomes expensive for this class of models as we increase $L$ and $L^\prime$---we would like to avoid deep GP models with a very large number of latent functions.}
We refer to this model as {\bf DGP}.

\subsubsection{Semi-Bayesian Gaussian Process Regression Networks}
\label{sec:sbgprn}

We now introduce the first model of interest in this work, namely a semi-Bayesian variant of
the model specified by Eqn.~\ref{eqn:gprn}. We simply `demote' $\bM(\bx_i)$ in Eqn.~\ref{eqn:gprn} to
a (deterministic) neural network:\footnote{For simplicity we regularize the neural network $\bM(\bx)$ with
L2-regularization on the weights, although other schemes are possible as well.}
\begin{equation}
\label{eqn:sbgprn}
\begin{split}
 P(\bY|\bX) \!=\! &\int \!  \Pi_{\ell=1}^L d\bff_\ell p(\bff_\ell | \bX)
 \prod_{i=1}^N  \NN(\by_i | \bM(\bx_i) \bF( \bx_i), \bbeta)
 \end{split}
\end{equation}
Below we refer to this model as {\bf SBGPRN}; it can be viewed as occupying an intermediate position between the
MOGP and GPRN.

\subsubsection{Neural Multi-Output Gaussian Processes}
\label{sec:nmogp}

A natural extension to the MOGP specified by Eqn.~\ref{eqn:mogp} is to pass the vector of Gaussian processes $\bF$ through
a layer of non-linearities before using $\bF$ to compute a mean function for the likelihood, i.e.~we
consider a model specified by its marginal likelihood as
\begin{equation}
\label{eqn:nmogp}
\int \! d\bM p(\bM) \prod_{\ell=1}^L d\bff_\ell p(\bff_\ell | \bX) \prod_{i=1}^N  \NN(\by_i | \bM \sigma(\bMt \bF( \bx_i)), \bbeta)
\end{equation}
where $\bMt$ is a $D_H \times D_L$ matrix\footnote{Throughout we treat $\bMt$ as a learnable parameter that is regularized via L2-regularization, i.e.~we place a Normal prior on $\bMt$ and perform MAP estimation on it.} and $\bM$ is a $D_Y \times D_H$ matrix, where
$D_H$ is a new hyperparameter that controls the number of `hidden units.' Here $\sigma(\cdot)$ is a fixed point-wise
non-linearity (e.g.~ReLU) and we place
a unit Normal prior on $\bM$. Below we refer to this model as {\bf N-MOGP}. Since this model does not contain
a (deterministic) neural network conditioned on the inputs as a subcomponent, we generally expect it to be less susceptible to overfitting than the SBGPRN. Note that, as is commonly done in the case of neural networks, we include a (stochastic) bias for each of the $D_H$ hidden units; see the supplementary materials for details.

\subsubsection{Neural Semi-Bayesian Gaussian Process Regression Networks}
\label{sec:nsbgprn}

In analogy to the Neural MOGP, a natural extension to the SBGPRN is to pass the vector of Gaussian processes $\bF$ through
a layer of non-linearities before applying the mixing matrix $\bM(\bx)$, yielding a model specified via its marginal probability as
\begin{equation}
\label{eqn:nsbgprn}
\begin{split}
\int \! \prod_{\ell=1}^L d\bff_\ell p(\bff_\ell | \bX)
\prod_{i=1}^N  \NN(\by_i | \bM(\bx_i)\sigma( \bMt \bF( \bx_i)), \bbeta)
 \end{split}
\end{equation}
Here $\sigma(\cdot)$ is a fixed non-linearity, $\bMt$ is a $D_H \times L$ matrix and $\bM(\bx)$ denotes a $D_Y \times D_H$ matrix controlled by a neural network. Here, again, $D_H$ is a hyperparameter that controls the number of `hidden units.'
Below we refer to this model as {\bf N-SBGPRN}.

\subsubsection{Neural Deep Gaussian Processes}
\label{sec:ndgp}

We equip the deep Gaussian process in Sec.~\ref{sec:dgp} with a neural likelihood:
\begin{equation}
\label{eqn:ndgp}
\begin{split}
 P(\bY|\bX) \!=\! \int \!  d\bM p(\bM) &\prod_{\ell=1}^L d\bff_\ell p(\bff_\ell | \bX)
 \!\prod_{\ell^\prime=1}^{L^\prime}\! d\bfft_{\ell^\prime} p(\bfft_{\ell^\prime} | \bff_{1:L})\\
 &\prod_{i=1}^N  \NN(\by_i | \bM \sigma(\bMt \bFt( \bx_i)), \bbeta)
 \end{split}
\end{equation}
where $\bMt$ and $\bM$ are $D_H \times L^\prime$ and $D_Y \times D_H$-sized matrices, respectively.
As above $p(\bM)$ denotes a unit Normal prior.
We refer to this model as {\bf N-DGP}.

\subsection{Models with Latent Inputs}
\label{sec:lvmmodels}

Each of the models in Sec.~\ref{sec:mogp}-\ref{sec:ndgp} can be repurposed as a model with latent inputs
by adding a prior on $\bX$. For example, for the MOGP in Sec.~\ref{sec:mogp} we have
\begin{equation}
\label{eqn:mogplvm}
\begin{split}
p(\bY ) = &\int d\bX d\bM p(\bX) p(\bM) \! \prod_{\ell=1}^L d\bff_\ell p(\bff_\ell | \bX) \times \\
&\prod_{i=1}^N  \NN(\by_i | \bM \bF( \bx_i), \bbeta)
\end{split}
\end{equation}
where $p(\bX)$ is a unit Normal prior on the inputs. We investigate a subset of these models empirically
in Sec.~\ref{sec:lvmexp}.

\section{Inference}
\label{sec:inf}

In this section we describe how we perform approximate inference for the models described in Sec.~\ref{sec:models}. In all cases we make use of variational inference due to its favorable computational properties and because it enables data subsampling during training.

\subsection{Sparse Variational Methods}
\label{sec:sparse}

In order to scale inference to large datasets we make use of sparse variational methods for Gaussian processes, which we now briefly review \citep{titsias2009variational, hensman2013gaussian}.  For every GP we introduce inducing
variables $\bu$ with ${\rm dim}(\bu)=N_{\rm ind}$, which are conditioned on $N_{\rm ind}$ variational parameters $\{ \bz_k \}$,\footnote{Unless noted otherwise, if there are multiple GPs we share the inducing points.}  with each $\bz_k$ of the
same dimension as the inputs to the GP. We then augment the GP prior with the inducing variables $\bu$
\begin{equation}\nonumber
p(\bff|\bX) \rightarrow p(\bff|\bu, \bX, \bZ) p(\bu | \bZ)
\end{equation}
and introduce a multivariate Normal variational distribution $q(\bu)$. We parameterize the covariance matrix
of $q(\bu)$ with a cholesky factor $\bm{L}$. For the variational distribution over $\bff$ we
choose the prior $p(\bff|\bu, \bX, \bZ)$ so that the variational distribution over
$(\bff, \bu)$ is $p(\bff|\bu, \bX, \bZ) q(\bu)$. By introducing the auxiliary variable $\bu$ we obtain a variational objective
that supports data subsampling, thus allowing us to scale to large datasets.
Before we discuss applying sparse methods to any particular model in Sec.~\ref{sec:models},
we first take a step back and discuss variational inference for models with Normal likelihoods.

\subsection{Variational Inference for Normal Likelihoods}
\label{sec:regvi}

We consider a regression model whose marginal likelihood is given by
\begin{equation}
\label{eqn:regressor}
p(\bY | \bX ) = \int \! d\bW p(\bW)  \prod_{i=1}^N  \NN(\by_i | \bPhi(\bx_i, \bW), \bbeta)
\end{equation}
where $\bPhi(\bx, \bW)$ is an arbitrary regressor function and $\bW$ denotes all the latent variables
in the model. Note that all the models in Sec.~\ref{sec:regmodels} can be expressed
in this form.\footnote{For example, for the MOGP in Eqn.~\ref{eqn:mogp} $\bW$ correspond to the $L$ latent
function values $\{ \bff_\ell \} = \{ \bF_i \}$ and the latent matrix $\bM$ and $\bPhi(\bx, \bW) = \bM \bF(\bx)$.} Introducing a variational distribution $q(\bW)$ the variational objective---i.e.~the evidence lower bound (ELBO)---can be written as
\begin{equation}
\label{eqn:regressorelbo}
\nonumber
{\rm ELBO} = \;\underbrace{\EE_{q(\bW)} \left[ \log p(\bY|\bX, \bW) \right]}_{{\rm ELL}} - {\rm KL}\!\left( q(\bW)||p(\bW)\right)
\end{equation}
where the first term is the expected log likelihood. We henceforth assume that the KL term is analytically tractable---as is
the case for all the models we consider---and focus on the expected log likelihood (ELL). At this point there are two possibilities:
i) we approximate the ELL with Monte Carlo samples; or ii) we compute the ELL analytically.
As we will make use of both approaches in our experiments, let us consider each possibility in turn.

\subsubsection{Stochastic Gradient Variational Bayes}
\label{sec:sgvb}

For all the models in Sec.~\ref{sec:models} we choose exclusively Normal variational distributions, which are amenable to the `reparameterization trick' \citep{price1958useful,kingma2013auto,rezende2014stochastic}. Consequently we can maximize
the ELBO using stochastic gradient methods.
At high level each iteration of training proceeds as follows:
\begin{enumerate}
\item subsample a mini-batch of data $(\bX_{\rm mb}, \bY_{\rm mb})$
\item form the variational distribution \\$q(\bff_{\rm mb}) \equiv \int d\bu p(\bff_{\rm mb}|\bu, \bX_{\rm mb}, \bZ) q(\bu)$
\item sample $\bff_{\rm mb} \sim q(\bff_{\rm mb})$ and compute a MC estimate of the expected log likelihood\\
         ${\rm ELL}_{\rm mb} = \EE_{q(\bff_{\rm mb})} \log p(\bY_{\rm mb} | \bff_{\rm mb})$
\item rescale  ${\rm ELL}_{\rm mb}$ to account for data subsampling
\item compute gradients of the ELBO with respect to model and variational parameters and take a gradient step
\end{enumerate}
We refer the reader to \citep{salimbeni2017doubly} for an application of stochastic gradient methods to the particular case of deep GPs.

\subsubsection{Analytic ELBOs}
\label{sec:analyticelbo}

For all the models in Sec.~\ref{sec:regmodels} apart from the deep Gaussian process models the expected log likelihood can either be computed analytically or---for those models with a non-linearity $\sigma$---almost analytically for a large class of non-linearities $\sigma$. Here by `almost' analytically we mean that everything can be computed analytically up to one-dimensional quadrature. We include a brief summary of this approach and refer the reader to the supplementary materials for details.

The expected log likelihood for a single datapoint $i$ can be rewritten as
\begin{equation}
\label{eqn:regressorell}
\begin{split}
&{\rm ELL}(i) = \;\EE_{q(\bW)} \left[ \log p(\by_i|\bX, \bW) \right] \\
                    &= \tfrac{1}{2}\sum_{k=1}^{D_Y} \log \tfrac{\beta_k}{2\pi} -
                     \sum_{k=1}^{D_Y} \tfrac{\beta_k}{2} \!\left \{ (y_{i,k}-\mean(\bx_i)_k)^2 + \var(\bx_i)_k\right\}
\end{split}
\end{equation}
where the $D_Y$-dimensional mean and variance functions $\bmean(\bx)$ and $\bvar(\bx)$ are defined as
\begin{equation}
\label{eqn:meanvar}
\begin{split}
\mean(\bx)_k = &\;\EE_{q(\bW)} \left[\Phi(\bx, \bW)_k \right] \\
\var(\bx)_k = &\;\EE_{q(\bW)} \left[ (\Phi(\bx, \bW)_k - \mean(\bx)_k)^2\right]
\end{split}
\end{equation}
For the MOGP and GPRN in Sec.~\ref{sec:mogp}-\ref{sec:gprn} as well as the SBGPRN in Sec.~\ref{sec:sbgprn} both of these quantities can be computed analytically.
For the N-MOGP and N-SBGPRN in Sec.~\ref{sec:nmogp}-\ref{sec:nsbgprn} the mean function
 $\bmean(\bx)$ can be computed analytically
for a wide class of non-linearities that includes, e.g., ReLU and the error function (erf).\footnote{More broadly, it includes
all piecewise polynomial non-linearities as well as non-linearities of the form $\sigma(x) = {\rm poly}(x) \erf(x)$, where
${\rm poly}(x)$ is polynomial.} Note that this implies that all of these models admit \emph{analytic} predictive means. For this same class of non-linearities the variance function $\bvar(\bx)$ can be reduced
to $\mathcal{O}(D_H^2)$ univariate Gaussian integrals, each of which can be efficiently computed using Gauss-Hermite quadrature; see the supplementary materials for details.

\subsubsubsection{{\bf Inference for the Neural MOGP}}
\label{sec:nmogpinf}

To make the proceeding overview more concrete, we provide a more detailed discussion of inference for
the Neural MOGP in \ref{sec:nmogp}, focusing on the case where the expected log likelihood
is computed analytically.

We place a diagonal Normal variational distribution $q(\bM)=\mathcal{N}(\bM|\bM_0, \bsig_\bM)$ on $\bM$.
We form $L$ multivariate Normal variational distributions
$\{ q(\bu_\ell) \}$ for the corresponding $L$ GPs.
Assuming we have analytic control over the non-linearity $\sigma$, we  compute the mean function $\bmean(\bx)$:
\begin{equation}
\label{eqn:nmogpmean}
\begin{split}
\mean(\bx_i)_k = &\;\EE_{q(\bM)\prod_\ell q(f_{\ell,i})} \left[ (\bM \sigma(\bMt \bF( \bx_i)))_k \right] \\
= &\; \Sigma_h M_{0, kh} \EE_{\prod_\ell q(f_{\ell,i})} \left[ (\sigma(\bMt \bF( \bx_i)))_h \right] \\
= &\; \Sigma_h M_{0, kh} \mean^\sigma_h(\bx_i)
\end{split}
\end{equation}
Here $q(f_{\ell,i}) = \int \! d\bu_\ell p(f_{\ell, i} | \bu_\ell, \bx_i, \bZ)  q(\bu_\ell)$ and
we have implicitly introduced the mean activation function $\bmean^\sigma(\bx)$ on the last line.
This quantity can be computed analytically as a function of $\bMt$
and the means and variances of the marginal (Normal) distributions $\{q(f_{\ell,i})\}$; see the supplementary materials for details.
Similarly we compute the variance function $\bvar(\bx)$:
\begin{equation}
\label{eqn:nmogpvar}
\begin{split}
\bvar(\bx) &= \bvar_1(\bx)  +\bvar_2(\bx)  +\bvar_3(\bx)  \qquad {\rm with} \\
\bvar_1(\bx)_k &\equiv  \Sigma_{h}\Sigma_{h^\prime}  M_{0, kh} \var^\sigma_{hh^\prime}(\bx) M_{0, kh^\prime} \\
\bvar_2(\bx)_k &\equiv \Sigma_h \sigma_{\bM,kh}^2 \mean^\sigma_h(\bx)^2 \\
\bvar_3(\bx)_k &\equiv \Sigma_h \sigma_{\bM,kh}^2 \var^\sigma_{hh}(\bx)  \\
\end{split}
\end{equation}
Here $\bvar^\sigma(\bx)$ is the $D_H \times D_H$ covariance matrix corresponding to $\bmean^\sigma(\bx)$.
This quantity can be computed efficiently using univariate quadrature, at a cost that scales quadratically in the number
of hidden units $D_H$; see the supplementary materials for details.

\subsection{Variational Inference for Models with Latent Inputs}
\label{sec:lvminf}

Inference for the models in Sec.~\ref{sec:lvmmodels} proceeds analogously to the models in Sec.~\ref{sec:regmodels},
with the difference that we now need to infer the latent inputs $\bX$. We introduce a factorized variational distribution
$q(\bX) = \prod_{i=1}^{N} q_i(\bx_i)$,
where each $q_i(\bx_i)$ is a Normal distribution with a diagonal covariance matrix.
During training we sample a mini-batch of latent inputs $\bX_{\rm mb} \sim q(\bX_{\rm mb})$
and make use of the reparameterization trick to compute gradients with respect to the variational parameters for the latent inputs.
Since we do not make use of an amortized variational distribution for the local latent variables $\{ \bx_i \}$, at test time we need to
fit a variational distribution $q(\bX^*)$ corresponding to test data $\bY^*$. For more details on the inference procedure, see the supplementary details.

\subsection{Fast Variational Inference for Sparse GPs}

The primary bottleneck for the inference procedures outlined above arises from dealing with the (potentially) large number of Gaussian processes. In particular, some of the most expensive subcomputations involved in computing the variational objective include:
\begin{enumerate}
\item computing KL divergences $\rm{KL}(q(\bu_\ell) | p(\bu_\ell))$
\item sampling from $q(\bff_\ell)$ when doing inference via SGVB as in Sec.~\ref{sec:sgvb}
\item computing the means and variances of the marginal distributions $q(f_{\ell, i})$ as required to compute analytic expected log likelihoods, c.f.~Sec.~\ref{sec:analyticelbo}
\end{enumerate}
For this reason we leverage modern conjugate gradient methods as implemented in GPyTorch \citep{gardner2018gpytorch}, which reduce the computational costs of 1-3 above from
$O(N_{\rm ind}^{3})$ to $O(N_{\rm ind}^{2})$.


%
%

\section{Experiments}
\label{sec:exp}
\newcommand{\llAfbax}{\tiny $13.16 \!\pm\! 0.46$}
\newcommand{\mrmseAfbax}{\tiny $0.285 \!\pm\! 0.004$}
\newcommand{\llAfbaxdk}{\tiny $14.02 \!\pm\! 0.80$}
\newcommand{\mrmseAfbaxdk}{\tiny $0.290 \!\pm\! 0.007$}
\newcommand{\llAkuka}{\tiny $15.64 \!\pm\! 0.40$}
\newcommand{\mrmseAkuka}{\tiny $0.174 \!\pm\! 0.006$}
\newcommand{\llAkukadk}{\tiny $16.32 \!\pm\! 1.06$}
\newcommand{\mrmseAkukadk}{\tiny $0.177 \!\pm\! 0.007$}
\newcommand{\llAmujo}{\tiny $-3.39 \!\pm\! 0.05$}
\newcommand{\mrmseAmujo}{\tiny $0.388 \!\pm\! 0.003$}
\newcommand{\llAmujodk}{\tiny $-1.97 \!\pm\! 0.13$}
\newcommand{\mrmseAmujodk}{\tiny $0.347 \!\pm\! 0.004$}
\newcommand{\llArbax}{\tiny $1.38 \!\pm\! 0.18$}
\newcommand{\mrmseArbax}{\tiny $0.297 \!\pm\! 0.007$}
\newcommand{\llArbaxdk}{\tiny $1.46 \!\pm\! 0.14$}
\newcommand{\mrmseArbaxdk}{\tiny $0.294 \!\pm\! 0.005$}
\newcommand{\llAsarc}{\tiny $1.63 \!\pm\! 0.04$}
\newcommand{\mrmseAsarc}{\tiny $0.250 \!\pm\! 0.001$}
\newcommand{\llAsarcdk}{\tiny $1.76 \!\pm\! 0.05$}
\newcommand{\mrmseAsarcdk}{\tiny $0.250 \!\pm\! 0.002$}
\newcommand{\llMxfbax}{\tiny $33.73 \!\pm\! 0.36$}
\newcommand{\mrmseMxfbax}{\tiny $0.040 \!\pm\! 0.001$}
\newcommand{\llMxfbaxdk}{\tiny $34.26 \!\pm\! 0.37$}
\newcommand{\mrmseMxfbaxdk}{\tiny $0.039 \!\pm\! 0.001$}
\newcommand{\llMxkuka}{\tiny $29.65 \!\pm\! 0.55$}
\newcommand{\mrmseMxkuka}{\tiny $0.087 \!\pm\! 0.001$}
\newcommand{\llMxkukadk}{\tiny $29.53 \!\pm\! 0.52$}
\newcommand{\mrmseMxkukadk}{\tiny $0.088 \!\pm\! 0.002$}
\newcommand{\llMxmujo}{\tiny $0.96 \!\pm\! 0.07$}
\newcommand{\mrmseMxmujo}{\tiny $0.231 \!\pm\! 0.002$}
\newcommand{\llMxmujodk}{\tiny $1.56 \!\pm\! 0.10$}
\newcommand{\mrmseMxmujodk}{\tiny $0.218 \!\pm\! 0.002$}
\newcommand{\llMxrbax}{\tiny $6.94 \!\pm\! 0.37$}
\newcommand{\mrmseMxrbax}{\tiny $0.116 \!\pm\! 0.006$}
\newcommand{\llMxrbaxdk}{\tiny $6.84 \!\pm\! 0.47$}
\newcommand{\mrmseMxrbaxdk}{\tiny $0.116 \!\pm\! 0.007$}
\newcommand{\llMxsarc}{\tiny $5.78 \!\pm\! 0.05$}
\newcommand{\mrmseMxsarc}{\tiny $0.113 \!\pm\! 0.001$}
\newcommand{\llMxsarcdk}{\tiny $5.93 \!\pm\! 0.06$}
\newcommand{\mrmseMxsarcdk}{\tiny $0.110 \!\pm\! 0.001$}


\newcommand{\llMsAfbaxerft}{\tiny $18.65 \!\pm\! 0.51$}
\newcommand{\mrmseMsAfbaxerft}{\tiny $0.124 \!\pm\! 0.016$}
\newcommand{\llMsAfbaxerftdk}{\tiny $19.14 \!\pm\! 0.55$}
\newcommand{\mrmseMsAfbaxerftdk}{\tiny $0.118 \!\pm\! 0.011$}

\newcommand{\llMsAfbaxerf}{\tiny $16.96 \!\pm\! 0.76$}
\newcommand{\mrmseMsAfbaxerf}{\tiny $0.177 \!\pm\! 0.031$}
\newcommand{\llMsAfbaxerfdk}{\tiny $17.49 \!\pm\! 0.47$}
\newcommand{\mrmseMsAfbaxerfdk}{\tiny $0.190 \!\pm\! 0.033$}

\newcommand{\llMsAfbaxleakyrelu}{\tiny $18.07 \!\pm\! 0.68$}
\newcommand{\mrmseMsAfbaxleakyrelu}{\tiny $0.136 \!\pm\! 0.016$}
\newcommand{\llMsAfbaxleakyreludk}{\tiny $18.43 \!\pm\! 0.57$}
\newcommand{\mrmseMsAfbaxleakyreludk}{\tiny $0.138 \!\pm\! 0.015$}

\newcommand{\llMsAfbaxrelu}{\tiny $12.23 \!\pm\! 2.73$}
\newcommand{\mrmseMsAfbaxrelu}{\tiny $0.196 \!\pm\! 0.039$}
\newcommand{\llMsAfbaxreludk}{\tiny $15.445 \!\pm\! 1.34$}
\newcommand{\mrmseMsAfbaxreludk}{\tiny $0.165 \!\pm\! 0.038$}

\newcommand{\llMsAkukaerft}{\tiny $23.61 \!\pm\! 0.35$}
\newcommand{\mrmseMsAkukaerft}{\tiny $0.118 \!\pm\! 0.014$}
\newcommand{\llMsAkukaerftdk}{\tiny $23.69 \!\pm\! 0.42$}
\newcommand{\mrmseMsAkukaerftdk}{\tiny $0.126 \!\pm\! 0.016$}

\newcommand{\llMsAkukaerf}{\tiny $22.42 \!\pm\! 0.53$}
\newcommand{\mrmseMsAkukaerf}{\tiny $0.126 \!\pm\! 0.015$}
\newcommand{\llMsAkukaerfdk}{\tiny $22.42 \!\pm\! 0.62$}
\newcommand{\mrmseMsAkukaerfdk}{\tiny $0.124 \!\pm\! 0.018$}

\newcommand{\llMsAkukaleakyrelu}{\tiny $22.07 \!\pm\! 0.59$}
\newcommand{\mrmseMsAkukaleakyrelu}{\tiny $0.114 \!\pm\! 0.011$}
\newcommand{\llMsAkukaleakyreludk}{\tiny $22.10 \!\pm\! 0.53$}
\newcommand{\mrmseMsAkukaleakyreludk}{\tiny $0.112 \!\pm\! 0.014$}

\newcommand{\llMsAkukarelu}{\tiny $19.68 \!\pm\! 1.49$}
\newcommand{\mrmseMsAkukarelu}{\tiny $0.146 \!\pm\! 0.013$}
\newcommand{\llMsAkukareludk}{\tiny $20.34 \!\pm\! 1.12$}
\newcommand{\mrmseMsAkukareludk}{\tiny $0.157 \!\pm\! 0.018$}

\newcommand{\llMsAmujoerft}{\tiny $-3.32 \!\pm\! 0.34$}
\newcommand{\mrmseMsAmujoerft}{\tiny $0.392 \!\pm\! 0.021$}
\newcommand{\llMsAmujoerftdk}{\tiny $-1.65 \!\pm\! 0.06$}
\newcommand{\mrmseMsAmujoerftdk}{\tiny $0.369 \!\pm\! 0.015$}

\newcommand{\llMsAmujoerf}{\tiny $-3.24 \!\pm\! 0.05$}
\newcommand{\mrmseMsAmujoerf}{\tiny $0.382 \!\pm\! 0.003$}
\newcommand{\llMsAmujoerfdk}{\tiny $-1.76 \!\pm\! 0.17$}
\newcommand{\mrmseMsAmujoerfdk}{\tiny $0.339 \!\pm\! 0.004$}

\newcommand{\llMsAmujoleakyrelu}{\tiny $-3.15 \!\pm\! 0.04$}
\newcommand{\mrmseMsAmujoleakyrelu}{\tiny $0.381 \!\pm\! 0.003$}
\newcommand{\llMsAmujoleakyreludk}{\tiny $-1.81 \!\pm\! 0.19$}
\newcommand{\mrmseMsAmujoleakyreludk}{\tiny $0.345 \!\pm\! 0.010$}

\newcommand{\llMsAmujorelu}{\tiny $-3.40 \!\pm\! 0.24$}
\newcommand{\mrmseMsAmujorelu}{\tiny $0.399 \!\pm\! 0.023$}
\newcommand{\llMsAmujoreludk}{\tiny $-1.96 \!\pm\! 0.21$}
\newcommand{\mrmseMsAmujoreludk}{\tiny $0.373 \!\pm\! 0.021$}

\newcommand{\llMsArbaxerft}{\tiny $3.81 \!\pm\! 0.22$}
\newcommand{\mrmseMsArbaxerft}{\tiny $0.228 \!\pm\! 0.007$}
\newcommand{\llMsArbaxerftdk}{\tiny $3.45 \!\pm\! 0.20$}
\newcommand{\mrmseMsArbaxerftdk}{\tiny $0.237 \!\pm\! 0.010$}

\newcommand{\llMsArbaxerf}{\tiny $3.78 \!\pm\! 0.37$}
\newcommand{\mrmseMsArbaxerf}{\tiny $0.239 \!\pm\! 0.027$}
\newcommand{\llMsArbaxerfdk}{\tiny $4.13 \!\pm\! 0.26$}
\newcommand{\mrmseMsArbaxerfdk}{\tiny $0.226 \!\pm\! 0.007$}

\newcommand{\llMsArbaxleakyrelu}{\tiny $3.53 \!\pm\! 0.21$}
\newcommand{\mrmseMsArbaxleakyrelu}{\tiny $0.228 \!\pm\! 0.009$}
\newcommand{\llMsArbaxleakyreludk}{\tiny $3.63 \!\pm\! 0.20$}
\newcommand{\mrmseMsArbaxleakyreludk}{\tiny $0.237 \!\pm\! 0.016$}

\newcommand{\llMsArbaxrelu}{\tiny $-3.42 \!\pm\! 2.63$}
\newcommand{\mrmseMsArbaxrelu}{\tiny $0.553 \!\pm\! 0.138$}
\newcommand{\llMsArbaxreludk}{\tiny $0.04 \!\pm\! 1.42$}
\newcommand{\mrmseMsArbaxreludk}{\tiny $0.377 \!\pm\! 0.080$}

\newcommand{\llMsAsarcerft}{\tiny $1.71 \!\pm\! 0.09$}
\newcommand{\mrmseMsAsarcerft}{\tiny $0.237 \!\pm\! 0.001$}
\newcommand{\llMsAsarcerftdk}{\tiny $1.92 \!\pm\! 0.13$}
\newcommand{\mrmseMsAsarcerftdk}{\tiny $0.240 \!\pm\! 0.009$}

\newcommand{\llMsAsarcerf}{\tiny $1.72 \!\pm\! 0.06$}
\newcommand{\mrmseMsAsarcerf}{\tiny $0.239 \!\pm\! 0.002$}
\newcommand{\llMsAsarcerfdk}{\tiny $1.95 \!\pm\! 0.05$}
\newcommand{\mrmseMsAsarcerfdk}{\tiny $0.242 \!\pm\! 0.006$}

\newcommand{\llMsAsarcleakyrelu}{\tiny $1.65 \!\pm\! 0.06$}
\newcommand{\mrmseMsAsarcleakyrelu}{\tiny $0.241 \!\pm\! 0.003$}
\newcommand{\llMsAsarcleakyreludk}{\tiny $1.96 \!\pm\! 0.19$}
\newcommand{\mrmseMsAsarcleakyreludk}{\tiny $0.236 \!\pm\! 0.015$}

\newcommand{\llMsAsarcrelu}{\tiny $1.66 \!\pm\! 0.14$}
\newcommand{\mrmseMsAsarcrelu}{\tiny $0.244 \!\pm\! 0.006$}
\newcommand{\llMsAsarcreludk}{\tiny $1.80 \!\pm\! 0.09$}
\newcommand{\mrmseMsAsarcreludk}{\tiny $0.246 \!\pm\! 0.011$}

\newcommand{\llMsMxfbaxerft}{\tiny $35.69 \!\pm\! 0.40$}
\newcommand{\mrmseMsMxfbaxerft}{\tiny $0.039 \!\pm\! 0.001$}
\newcommand{\llMsMxfbaxerftdk}{\tiny $36.24 \!\pm\! 0.36$}
\newcommand{\mrmseMsMxfbaxerftdk}{\tiny $0.038 \!\pm\! 0.001$}

\newcommand{\llMsMxfbaxerf}{\tiny $34.59 \!\pm\! 0.33$}
\newcommand{\mrmseMsMxfbaxerf}{\tiny $0.042 \!\pm\! 0.001$}
\newcommand{\llMsMxfbaxerfdk}{\tiny $35.41 \!\pm\! 0.32$}
\newcommand{\mrmseMsMxfbaxerfdk}{\tiny $0.041 \!\pm\! 0.001$}

\newcommand{\llMsMxfbaxleakyrelu}{\tiny $35.23 \!\pm\! 0.33$}
\newcommand{\mrmseMsMxfbaxleakyrelu}{\tiny $0.039 \!\pm\! 0.001$}
\newcommand{\llMsMxfbaxleakyreludk}{\tiny $36.01 \!\pm\! 0.34$}
\newcommand{\mrmseMsMxfbaxleakyreludk}{\tiny $0.038 \!\pm\! 0.001$}

\newcommand{\llMsMxfbaxrelu}{\tiny $35.75 \!\pm\! 0.26$}
\newcommand{\mrmseMsMxfbaxrelu}{\tiny $0.038 \!\pm\! 0.001$}
\newcommand{\llMsMxfbaxreludk}{\tiny $36.66 \!\pm\! 0.30$}
\newcommand{\mrmseMsMxfbaxreludk}{\tiny $0.037 \!\pm\! 0.001$}

\newcommand{\llMsMxkukaerft}{\tiny $31.70 \!\pm\! 0.36$}
\newcommand{\mrmseMsMxkukaerft}{\tiny $0.087 \!\pm\! 0.002$}
\newcommand{\llMsMxkukaerftdk}{\tiny $32.30 \!\pm\! 0.48$}
\newcommand{\mrmseMsMxkukaerftdk}{\tiny $0.086 \!\pm\! 0.001$}

\newcommand{\llMsMxkukaerf}{\tiny $29.83 \!\pm\! 0.63$}
\newcommand{\mrmseMsMxkukaerf}{\tiny $0.088 \!\pm\! 0.001$}
\newcommand{\llMsMxkukaerfdk}{\tiny $30.81 \!\pm\! 0.54$}
\newcommand{\mrmseMsMxkukaerfdk}{\tiny $0.088 \!\pm\! 0.001$}

\newcommand{\llMsMxkukaleakyrelu}{\tiny $31.19 \!\pm\! 0.33$}
\newcommand{\mrmseMsMxkukaleakyrelu}{\tiny $0.087 \!\pm\! 0.001$}
\newcommand{\llMsMxkukaleakyreludk}{\tiny $31.58 \!\pm\! 0.39$}
\newcommand{\mrmseMsMxkukaleakyreludk}{\tiny $0.086 \!\pm\! 0.001$}

\newcommand{\llMsMxkukarelu}{\tiny $31.30 \!\pm\! 0.46$}
\newcommand{\mrmseMsMxkukarelu}{\tiny $0.087 \!\pm\! 0.002$}
\newcommand{\llMsMxkukareludk}{\tiny $31.91 \!\pm\! 0.44$}
\newcommand{\mrmseMsMxkukareludk}{\tiny $0.086 \!\pm\! 0.001$}

\newcommand{\llMsMxmujoerft}{\tiny $1.24 \!\pm\! 0.10$}
\newcommand{\mrmseMsMxmujoerft}{\tiny $0.226 \!\pm\! 0.002$}
\newcommand{\llMsMxmujoerftdk}{\tiny $1.60 \!\pm\! 0.10$}
\newcommand{\mrmseMsMxmujoerftdk}{\tiny $0.217 \!\pm\! 0.003$}

\newcommand{\llMsMxmujoerf}{\tiny $1.06 \!\pm\! 0.10$}
\newcommand{\mrmseMsMxmujoerf}{\tiny $0.231 \!\pm\! 0.003$}
\newcommand{\llMsMxmujoerfdk}{\tiny $1.41 \!\pm\! 0.08$}
\newcommand{\mrmseMsMxmujoerfdk}{\tiny $0.222 \!\pm\! 0.002$}

\newcommand{\llMsMxmujoleakyrelu}{\tiny $1.54 \!\pm\! 0.08$}
\newcommand{\mrmseMsMxmujoleakyrelu}{\tiny $0.219 \!\pm\! 0.002$}
\newcommand{\llMsMxmujoleakyreludk}{\tiny $1.87 \!\pm\! 0.12$}
\newcommand{\mrmseMsMxmujoleakyreludk}{\tiny $0.212 \!\pm\! 0.003$}

\newcommand{\llMsMxmujorelu}{\tiny $1.52 \!\pm\! 0.08$}
\newcommand{\mrmseMsMxmujorelu}{\tiny $0.220 \!\pm\! 0.003$}
\newcommand{\llMsMxmujoreludk}{\tiny $1.86 \!\pm\! 0.08$}
\newcommand{\mrmseMsMxmujoreludk}{\tiny $0.211 \!\pm\! 0.002$}

\newcommand{\llMsMxrbaxerft}{\tiny $6.72 \!\pm\! 0.16$}
\newcommand{\mrmseMsMxrbaxerft}{\tiny $0.119 \!\pm\! 0.002$}
\newcommand{\llMsMxrbaxerftdk}{\tiny $6.66 \!\pm\! 0.40$}
\newcommand{\mrmseMsMxrbaxerftdk}{\tiny $0.118 \!\pm\! 0.008$}

\newcommand{\llMsMxrbaxerf}{\tiny $7.47 \!\pm\! 0.26$}
\newcommand{\mrmseMsMxrbaxerf}{\tiny $0.105 \!\pm\! 0.004$}
\newcommand{\llMsMxrbaxerfdk}{\tiny $7.55 \!\pm\! 0.21$}
\newcommand{\mrmseMsMxrbaxerfdk}{\tiny $0.107 \!\pm\! 0.003$}

\newcommand{\llMsMxrbaxleakyrelu}{\tiny $7.55 \!\pm\! 0.18$}
\newcommand{\mrmseMsMxrbaxleakyrelu}{\tiny $0.105 \!\pm\! 0.002$}
\newcommand{\llMsMxrbaxleakyreludk}{\tiny $7.80 \!\pm\! 0.14$}
\newcommand{\mrmseMsMxrbaxleakyreludk}{\tiny $0.107 \!\pm\! 0.003$}

\newcommand{\llMsMxrbaxrelu}{\tiny $7.24 \!\pm\! 0.35$}
\newcommand{\mrmseMsMxrbaxrelu}{\tiny $0.110 \!\pm\! 0.006$}
\newcommand{\llMsMxrbaxreludk}{\tiny $7.52 \!\pm\! 0.29$}
\newcommand{\mrmseMsMxrbaxreludk}{\tiny $0.104 \!\pm\! 0.003$}

\newcommand{\llMsMxsarcerft}{\tiny $6.00 \!\pm\! 0.07$}
\newcommand{\mrmseMsMxsarcerft}{\tiny $0.110 \!\pm\! 0.001$}
\newcommand{\llMsMxsarcerftdk}{\tiny $6.13 \!\pm\! 0.06$}
\newcommand{\mrmseMsMxsarcerftdk}{\tiny $0.107 \!\pm\! 0.001$}

\newcommand{\llMsMxsarcerf}{\tiny $5.80 \!\pm\! 0.07$}
\newcommand{\mrmseMsMxsarcerf}{\tiny $0.113 \!\pm\! 0.001$}
\newcommand{\llMsMxsarcerfdk}{\tiny $5.93 \!\pm\! 0.10$}
\newcommand{\mrmseMsMxsarcerfdk}{\tiny $0.110 \!\pm\! 0.001$}

\newcommand{\llMsMxsarcleakyrelu}{\tiny $6.02 \!\pm\! 0.07$}
\newcommand{\mrmseMsMxsarcleakyrelu}{\tiny $0.109 \!\pm\! 0.001$}
\newcommand{\llMsMxsarcleakyreludk}{\tiny $6.14 \!\pm\! 0.09$}
\newcommand{\mrmseMsMxsarcleakyreludk}{\tiny $0.107 \!\pm\! 0.001$}

\newcommand{\llMsMxsarcrelu}{\tiny $5.90 \!\pm\! 0.13$}
\newcommand{\mrmseMsMxsarcrelu}{\tiny $0.111 \!\pm\! 0.002$}
\newcommand{\llMsMxsarcreludk}{\tiny $6.04 \!\pm\! 0.10$}
\newcommand{\mrmseMsMxsarcreludk}{\tiny $0.108 \!\pm\! 0.001$}

\newcommand{\llMxsAfbaxerft}{\tiny $30.8198 \!\pm\! 0.7983$}
\newcommand{\mrmseMxsAfbaxerft}{\tiny $0.054 \!\pm\! 0.002$}
\newcommand{\llMxsAfbaxerftdk}{\tiny $31.12 \!\pm\! 1.15$}
\newcommand{\mrmseMxsAfbaxerftdk}{\tiny $0.054 \!\pm\! 0.005$}

\newcommand{\llMxsAfbaxerf}{\tiny $33.91 \!\pm\! 0.38$}
\newcommand{\mrmseMxsAfbaxerf}{\tiny $0.043 \!\pm\! 0.002$}
\newcommand{\llMxsAfbaxerfdk}{\tiny $34.91 \!\pm\! 0.37$}
\newcommand{\mrmseMxsAfbaxerfdk}{\tiny $0.041 \!\pm\! 0.002$}

\newcommand{\llMxsAfbaxleakyrelu}{\tiny $31.41 \!\pm\! 0.14$}
\newcommand{\mrmseMxsAfbaxleakyrelu}{\tiny $0.052 \!\pm\! 0.001$}
\newcommand{\llMxsAfbaxleakyreludk}{\tiny $31.92 \!\pm\! 0.27$}
\newcommand{\mrmseMxsAfbaxleakyreludk}{\tiny $0.051 \!\pm\! 0.001$}

\newcommand{\llMxsAfbaxrelu}{\tiny $29.73 \!\pm\! 1.06$}
\newcommand{\mrmseMxsAfbaxrelu}{\tiny $0.062 \!\pm\! 0.008$}
\newcommand{\llMxsAfbaxreludk}{\tiny $30.10 \!\pm\! 1.28$}
\newcommand{\mrmseMxsAfbaxreludk}{\tiny $0.059 \!\pm\! 0.005$}

\newcommand{\llMxsAkukaerft}{\tiny $27.98 \!\pm\! 0.52$}
\newcommand{\mrmseMxsAkukaerft}{\tiny $0.091 \!\pm\! 0.002$}
\newcommand{\llMxsAkukaerftdk}{\tiny $28.28 \!\pm\! 0.30$}
\newcommand{\mrmseMxsAkukaerftdk}{\tiny $0.091 \!\pm\! 0.001$}

\newcommand{\llMxsAkukaerf}{\tiny $30.15 \!\pm\! 0.34$}
\newcommand{\mrmseMxsAkukaerf}{\tiny $0.087 \!\pm\! 0.002$}
\newcommand{\llMxsAkukaerfdk}{\tiny $30.87 \!\pm\! 0.47$}
\newcommand{\mrmseMxsAkukaerfdk}{\tiny $0.087 \!\pm\! 0.001$}

\newcommand{\llMxsAkukaleakyrelu}{\tiny $27.92 \!\pm\! 0.44$}
\newcommand{\mrmseMxsAkukaleakyrelu}{\tiny $0.090 \!\pm\! 0.001$}
\newcommand{\llMxsAkukaleakyreludk}{\tiny $27.94 \!\pm\! 0.45$}
\newcommand{\mrmseMxsAkukaleakyreludk}{\tiny $0.090 \!\pm\! 0.002$}

\newcommand{\llMxsAkukarelu}{\tiny $26.47 \!\pm\! 0.77$}
\newcommand{\mrmseMxsAkukarelu}{\tiny $0.095 \!\pm\! 0.003$}
\newcommand{\llMxsAkukareludk}{\tiny $27.00 \!\pm\! 0.52$}
\newcommand{\mrmseMxsAkukareludk}{\tiny $0.095 \!\pm\! 0.004$}

\newcommand{\llMxsAmujoerft}{\tiny $0.82 \!\pm\! 0.10$}
\newcommand{\mrmseMxsAmujoerft}{\tiny $0.232 \!\pm\! 0.003$}
\newcommand{\llMxsAmujoerftdk}{\tiny $0.82 \!\pm\! 0.16$}
\newcommand{\mrmseMxsAmujoerftdk}{\tiny $0.234 \!\pm\! 0.005$}

\newcommand{\llMxsAmujoerf}{\tiny $0.74 \!\pm\! 0.22$}
\newcommand{\mrmseMxsAmujoerf}{\tiny $0.235 \!\pm\! 0.005$}
\newcommand{\llMxsAmujoerfdk}{\tiny $1.259 \!\pm\! 0.131$}
\newcommand{\mrmseMxsAmujoerfdk}{\tiny $0.225 \!\pm\! 0.002$}

\newcommand{\llMxsAmujoleakyrelu}{\tiny $0.84 \!\pm\! 0.07$}
\newcommand{\mrmseMxsAmujoleakyrelu}{\tiny $0.231 \!\pm\! 0.002$}
\newcommand{\llMxsAmujoleakyreludk}{\tiny $1.48 \!\pm\! 0.06$}
\newcommand{\mrmseMxsAmujoleakyreludk}{\tiny $0.217 \!\pm\! 0.002$}

\newcommand{\llMxsAmujorelu}{\tiny $0.87 \!\pm\! 0.11$}
\newcommand{\mrmseMxsAmujorelu}{\tiny $0.230 \!\pm\! 0.003$}
\newcommand{\llMxsAmujoreludk}{\tiny $1.45 \!\pm\! 0.16$}
\newcommand{\mrmseMxsAmujoreludk}{\tiny $0.217 \!\pm\! 0.003$}

\newcommand{\llMxsArbaxerft}{\tiny $-6.86 \!\pm\! 6.18$}
\newcommand{\mrmseMxsArbaxerft}{\tiny $0.845 \!\pm\! 0.315$}
\newcommand{\llMxsArbaxerftdk}{\tiny $2.47 \!\pm\! 6.24$}
\newcommand{\mrmseMxsArbaxerftdk}{\tiny $0.367 \!\pm\! 0.316$}

\newcommand{\llMxsArbaxerf}{\tiny $-9.95 \!\pm\! 0.08$}
\newcommand{\mrmseMxsArbaxerf}{\tiny $1.002 \!\pm\! 0.010$}
\newcommand{\llMxsArbaxerfdk}{\tiny $-9.95 \!\pm\! 0.07$}
\newcommand{\mrmseMxsArbaxerfdk}{\tiny $1.002 \!\pm\! 0.010$}

\newcommand{\llMxsArbaxleakyrelu}{\tiny $-9.95 \!\pm\! 0.07$}
\newcommand{\mrmseMxsArbaxleakyrelu}{\tiny $1.002 \!\pm\! 0.010$}
\newcommand{\llMxsArbaxleakyreludk}{\tiny $-8.45 \!\pm\! 4.49$}
\newcommand{\mrmseMxsArbaxleakyreludk}{\tiny $0.923 \!\pm\! 0.237$}

\newcommand{\llMxsArbaxrelu}{\tiny $-7.93 \!\pm\! 3.31$}
\newcommand{\mrmseMxsArbaxrelu}{\tiny $0.865 \!\pm\! 0.214$}
\newcommand{\llMxsArbaxreludk}{\tiny $-7.69 \!\pm\! 3.77$}
\newcommand{\mrmseMxsArbaxreludk}{\tiny $0.844 \!\pm\! 0.253$}

\newcommand{\llMxsAsarcerft}{\tiny $4.98 \!\pm\! 0.04$}
\newcommand{\mrmseMxsAsarcerft}{\tiny $0.127 \!\pm\! 0.001$}
\newcommand{\llMxsAsarcerftdk}{\tiny $4.95 \!\pm\! 0.07$}
\newcommand{\mrmseMxsAsarcerftdk}{\tiny $0.128 \!\pm\! 0.001$}

\newcommand{\llMxsAsarcerf}{\tiny $5.88 \!\pm\! 0.08$}
\newcommand{\mrmseMxsAsarcerf}{\tiny $0.111 \!\pm\! 0.001$}
\newcommand{\llMxsAsarcerfdk}{\tiny $5.91 \!\pm\! 0.07$}
\newcommand{\mrmseMxsAsarcerfdk}{\tiny $0.111 \!\pm\! 0.001$}

\newcommand{\llMxsAsarcleakyrelu}{\tiny $5.06 \!\pm\! 0.15$}
\newcommand{\mrmseMxsAsarcleakyrelu}{\tiny $0.126 \!\pm\! 0.003$}
\newcommand{\llMxsAsarcleakyreludk}{\tiny $5.19 \!\pm\! 0.27$}
\newcommand{\mrmseMxsAsarcleakyreludk}{\tiny $0.123 \!\pm\! 0.005$}

\newcommand{\llMxsAsarcrelu}{\tiny $5.04 \!\pm\! 0.04$}
\newcommand{\mrmseMxsAsarcrelu}{\tiny $0.126 \!\pm\! 0.001$}
\newcommand{\llMxsAsarcreludk}{\tiny $5.12 \!\pm\! 0.06$}
\newcommand{\mrmseMxsAsarcreludk}{\tiny $0.125 \!\pm\! 0.001$}


\newcommand{\llgprnfbax}{\tiny $23.74 \!\pm\! 0.79$}
\newcommand{\mrmsegprnfbax}{\tiny $0.042 \!\pm\! 0.002$}
\newcommand{\llgprnfbaxdk}{\tiny $24.77 \!\pm\! 0.79$}
\newcommand{\mrmsegprnfbaxdk}{\tiny $0.040 \!\pm\! 0.001$}
\newcommand{\llgprnkuka}{\tiny $17.57 \!\pm\! 0.64$}
\newcommand{\mrmsegprnkuka}{\tiny $0.091 \!\pm\! 0.002$}
\newcommand{\llgprnkukadk}{\tiny $19.14 \!\pm\! 0.64$}
\newcommand{\mrmsegprnkukadk}{\tiny $0.089 \!\pm\! 0.002$}
\newcommand{\llgprnmujo}{\tiny $-2.50 \!\pm\! 0.11$}
\newcommand{\mrmsegprnmujo}{\tiny $0.336 \!\pm\! 0.004$}
\newcommand{\llgprnmujodk}{\tiny $-1.83 \!\pm\! 0.35$}
\newcommand{\mrmsegprnmujodk}{\tiny $0.315 \!\pm\! 0.011$}
\newcommand{\llgprnrbax}{\tiny $3.80 \!\pm\! 0.22$}
\newcommand{\mrmsegprnrbax}{\tiny $0.144 \!\pm\! 0.007$}
\newcommand{\llgprnrbaxdk}{\tiny $2.78 \!\pm\! 0.38$}
\newcommand{\mrmsegprnrbaxdk}{\tiny $0.179 \!\pm\! 0.009$}
\newcommand{\llgprnsarc}{\tiny $4.79 \!\pm\! 0.05$}
\newcommand{\mrmsegprnsarc}{\tiny $0.125 \!\pm\! 0.001$}
\newcommand{\llgprnsarcdk}{\tiny $4.52 \!\pm\! 0.11$}
\newcommand{\mrmsegprnsarcdk}{\tiny $0.133 \!\pm\! 0.003$}


\newcommand{\llMsAfbaxerftdh}{\tiny $25.50 \!\pm\! 0.24$}
\newcommand{\mrmseMsAfbaxerftdh}{\tiny $0.068 \!\pm\! 0.001$}
\newcommand{\llMsAfbaxerfdh}{\tiny $18.37 \!\pm\! 0.77$}
\newcommand{\mrmseMsAfbaxerfdh}{\tiny $0.128 \!\pm\! 0.016$}
\newcommand{\llMsAfbaxleakyreludh}{\tiny $23.00 \!\pm\! 0.50$}
\newcommand{\mrmseMsAfbaxleakyreludh}{\tiny $0.075 \!\pm\! 0.002$}
\newcommand{\llMsAfbaxreludh}{\tiny $18.53 \!\pm\! 0.61$}
\newcommand{\mrmseMsAfbaxreludh}{\tiny $0.103 \!\pm\! 0.011$}
\newcommand{\llMsAkukaerftdh}{\tiny $25.32 \!\pm\! 0.26$}
\newcommand{\mrmseMsAkukaerftdh}{\tiny $0.095 \!\pm\! 0.002$}
\newcommand{\llMsAkukaerfdh}{\tiny $22.90 \!\pm\! 0.43$}
\newcommand{\mrmseMsAkukaerfdh}{\tiny $0.112 \!\pm\! 0.006$}
\newcommand{\llMsAkukaleakyreludh}{\tiny $23.50 \!\pm\! 0.48$}
\newcommand{\mrmseMsAkukaleakyreludh}{\tiny $0.097 \!\pm\! 0.002$}
\newcommand{\llMsAkukareludh}{\tiny $21.64 \!\pm\! 0.79$}
\newcommand{\mrmseMsAkukareludh}{\tiny $0.107 \!\pm\! 0.007$}
\newcommand{\llMsAmujoerftdh}{\tiny $-3.06 \!\pm\! 0.17$}
\newcommand{\mrmseMsAmujoerftdh}{\tiny $0.383 \!\pm\! 0.014$}
\newcommand{\llMsAmujoerfdh}{\tiny $-3.20 \!\pm\! 0.33$}
\newcommand{\mrmseMsAmujoerfdh}{\tiny $0.384 \!\pm\! 0.021$}
\newcommand{\llMsAmujoleakyreludh}{\tiny $-2.92 \!\pm\! 0.05$}
\newcommand{\mrmseMsAmujoleakyreludh}{\tiny $0.373 \!\pm\! 0.003$}
\newcommand{\llMsAmujoreludh}{\tiny $-3.02 \!\pm\! 0.05$}
\newcommand{\mrmseMsAmujoreludh}{\tiny $0.377 \!\pm\! 0.003$}
\newcommand{\llMsArbaxerftdh}{\tiny $4.83 \!\pm\! 0.22$}
\newcommand{\mrmseMsArbaxerftdh}{\tiny $0.186 \!\pm\! 0.018$}
\newcommand{\llMsArbaxerfdh}{\tiny $4.59 \!\pm\! 0.55$}
\newcommand{\mrmseMsArbaxerfdh}{\tiny $0.183 \!\pm\! 0.028$}
\newcommand{\llMsArbaxleakyreludh}{\tiny $4.80 \!\pm\! 0.34$}
\newcommand{\mrmseMsArbaxleakyreludh}{\tiny $0.174 \!\pm\! 0.016$}
\newcommand{\llMsArbaxreludh}{\tiny $0.88 \!\pm\! 1.76$}
\newcommand{\mrmseMsArbaxreludh}{\tiny $0.365 \!\pm\! 0.089$}
\newcommand{\llMsAsarcerftdh}{\tiny $2.43 \!\pm\! 0.21$}
\newcommand{\mrmseMsAsarcerftdh}{\tiny $0.194 \!\pm\! 0.011$}
\newcommand{\llMsAsarcerfdh}{\tiny $2.11 \!\pm\! 0.25$}
\newcommand{\mrmseMsAsarcerfdh}{\tiny $0.218 \!\pm\! 0.015$}
\newcommand{\llMsAsarcleakyreludh}{\tiny $2.16 \!\pm\! 0.15$}
\newcommand{\mrmseMsAsarcleakyreludh}{\tiny $0.212 \!\pm\! 0.008$}
\newcommand{\llMsAsarcreludh}{\tiny $1.99 \!\pm\! 0.10$}
\newcommand{\mrmseMsAsarcreludh}{\tiny $0.220 \!\pm\! 0.007$}


\newcommand{\lldeepgprbax}{\tiny $6.34 \!\pm\! 0.12$}
\newcommand{\mrmsedeepgprbax}{\tiny $0.200 \!\pm\! 0.004$}

\newcommand{\lldeepgpkuka}{\tiny $25.08 \!\pm\! 0.22$}
\newcommand{\mrmsedeepgpkuka}{\tiny $0.089 \!\pm\! 0.001$}

\newcommand{\lldeepgpfbax}{\tiny $23.81 \!\pm\! 0.31$}
\newcommand{\mrmsedeepgpfbax}{\tiny $0.083 \!\pm\! 0.001$}

\newcommand{\lldeepgpsarc}{\tiny $3.45 \!\pm\! 0.11$}
\newcommand{\mrmsedeepgpsarc}{\tiny $0.166 \!\pm\! 0.003$}

\newcommand{\lldeepgpmujo}{\tiny $-2.46 \!\pm\! 0.16$}
\newcommand{\mrmsedeepgpmujo}{\tiny $0.354 \!\pm\! 0.013$}


\newcommand{\llndgprbaxerftdh}{\tiny $3.27 \!\pm\! 2.58$}
\newcommand{\mrmsendgprbaxerftdh}{\tiny $0.232 \!\pm\! 0.082$}

\newcommand{\llndgpkukaerftdh}{\tiny $25.63 \!\pm\! 0.47$}
\newcommand{\mrmsendgpkukaerftdh}{\tiny $0.091 \!\pm\! 0.002$}

\newcommand{\llndgpfbaxerftdh}{\tiny $27.42 \!\pm\! 0.49$}
\newcommand{\mrmsendgpfbaxerftdh}{\tiny $0.058 \!\pm\! 0.003$}

\newcommand{\llndgpsarcerftdh}{\tiny $2.95 \!\pm\! 0.82$}
\newcommand{\mrmsendgpsarcerftdh}{\tiny $0.192 \!\pm\! 0.035$}

\newcommand{\llndgpmujoerftdh}{\tiny $-2.41 \!\pm\! 0.12$}
\newcommand{\mrmsendgpmujoerftdh}{\tiny $0.357 \!\pm\! 0.009$}

\newcommand{\llndgprbaxerfdh}{\tiny $7.30 \!\pm\! 0.23$}
\newcommand{\mrmsendgprbaxerfdh}{\tiny $0.110 \!\pm\! 0.004$}

\newcommand{\llndgpkukaerfdh}{\tiny $25.41 \!\pm\! 0.37$}
\newcommand{\mrmsendgpkukaerfdh}{\tiny $0.088 \!\pm\! 0.001$}

\newcommand{\llndgpfbaxerfdh}{\tiny $27.36 \!\pm\! 0.42$}
\newcommand{\mrmsendgpfbaxerfdh}{\tiny $0.057 \!\pm\! 0.002$}

\newcommand{\llndgpsarcerfdh}{\tiny $3.69 \!\pm\! 0.07$}
\newcommand{\mrmsendgpsarcerfdh}{\tiny $0.158 \!\pm\! 0.003$}

\newcommand{\llndgpmujoerfdh}{\tiny $-2.41 \!\pm\! 0.15$}
\newcommand{\mrmsendgpmujoerfdh}{\tiny $0.348 \!\pm\! 0.009$}

\newcommand{\llndgprbaxleakyreludh}{\tiny $7.42 \!\pm\! 0.17$}
\newcommand{\mrmsendgprbaxleakyreludh}{\tiny $0.107 \!\pm\! 0.003$}

\newcommand{\llndgpkukaleakyreludh}{\tiny $25.66 \!\pm\! 0.29$}
\newcommand{\mrmsendgpkukaleakyreludh}{\tiny $0.089 \!\pm\! 0.001$}

\newcommand{\llndgpfbaxleakyreludh}{\tiny $27.93 \!\pm\! 0.21$}
\newcommand{\mrmsendgpfbaxleakyreludh}{\tiny $0.053 \!\pm\! 0.001$}

\newcommand{\llndgpsarcleakyreludh}{\tiny $3.65 \!\pm\! 0.10$}
\newcommand{\mrmsendgpsarcleakyreludh}{\tiny $0.161 \!\pm\! 0.005$}

\newcommand{\llndgpmujoleakyreludh}{\tiny $-2.99 \!\pm\! 0.96$}
\newcommand{\mrmsendgpmujoleakyreludh}{\tiny $0.390 \!\pm\! 0.052$}


\newcommand{\lldeepgprbaxsmallL}{\tiny $2.91 \pm 0.29$}
\newcommand{\mrmsedeepgprbaxsmallL}{\tiny $0.315 \pm 0.029$}

\newcommand{\lldeepgpkukasmallL}{\tiny $15.60 \pm 0.36$}
\newcommand{\mrmsedeepgpkukasmallL}{\tiny $0.190 \pm 0.007$}

\newcommand{\lldeepgpfbaxsmallL}{\tiny $12.89 \pm 0.69$}
\newcommand{\mrmsedeepgpfbaxsmallL}{\tiny $0.284 \pm 0.005$}

\newcommand{\lldeepgpsarcsmallL}{\tiny $2.31 \pm 0.05$}
\newcommand{\mrmsedeepgpsarcsmallL}{\tiny $0.243 \pm 0.002$}

\newcommand{\lldeepgpmujosmallL}{\tiny $-2.81 \pm 0.16$}
\newcommand{\mrmsedeepgpmujosmallL}{\tiny $0.384 \pm 0.021$}

\newcommand{\llndgprbaxerftsmallL}{\tiny $3.21 \pm 2.51$}
\newcommand{\mrmsendgprbaxerftsmallL}{\tiny $0.238 \pm 0.084$}

\newcommand{\llndgpkukaerftsmallL}{\tiny $24.73 \pm 0.46$}
\newcommand{\mrmsendgpkukaerftsmallL}{\tiny $0.093 \pm 0.002$}

\newcommand{\llndgpfbaxerftsmallL}{\tiny $24.63 \pm 0.97$}
\newcommand{\mrmsendgpfbaxerftsmallL}{\tiny $0.068 \pm 0.004$}

\newcommand{\llndgpsarcerftsmallL}{\tiny $2.93 \pm 0.21$}
\newcommand{\mrmsendgpsarcerftsmallL}{\tiny $0.201 \pm 0.010$}

\newcommand{\llndgpmujoerftsmallL}{\tiny $-2.68 \pm 0.17$}
\newcommand{\mrmsendgpmujoerftsmallL}{\tiny $0.374 \pm 0.018$}

\newcommand{\llndgprbaxleakyrelusmallL}{\tiny $5.83 \pm 0.18$}
\newcommand{\mrmsendgprbaxleakyrelusmallL}{\tiny $0.169 \pm 0.016$}

\newcommand{\llndgpkukaleakyrelusmallL}{\tiny $24.61 \pm 0.32$}
\newcommand{\mrmsendgpkukaleakyrelusmallL}{\tiny $0.092 \pm 0.001$}

\newcommand{\llndgpfbaxleakyrelusmallL}{\tiny $24.43 \pm 0.41$}
\newcommand{\mrmsendgpfbaxleakyrelusmallL}{\tiny $0.067 \pm 0.001$}

\newcommand{\llndgpsarcleakyrelusmallL}{\tiny $2.97 \pm 0.24$}
\newcommand{\mrmsendgpsarcleakyrelusmallL}{\tiny $0.197 \pm 0.012$}

\newcommand{\llndgpmujoleakyrelusmallL}{\tiny $-3.26 \pm 0.54$}
\newcommand{\mrmsendgpmujoleakyrelusmallL}{\tiny $0.405 \pm 0.027$}

\newcommand{\llndgprbaxerfsmallL}{\tiny $5.98 \pm 0.20$}
\newcommand{\mrmsendgprbaxerfsmallL}{\tiny $0.145 \pm 0.017$}

\newcommand{\llndgpkukaerfsmallL}{\tiny $23.69 \pm 0.23$}
\newcommand{\mrmsendgpkukaerfsmallL}{\tiny $0.103 \pm 0.004$}

\newcommand{\llndgpfbaxerfsmallL}{\tiny $21.77 \pm 0.65$}
\newcommand{\mrmsendgpfbaxerfsmallL}{\tiny $0.081 \pm 0.005$}

\newcommand{\llndgpsarcerfsmallL}{\tiny $2.82 \pm 0.16$}
\newcommand{\mrmsendgpsarcerfsmallL}{\tiny $0.214 \pm 0.010$}

\newcommand{\llndgpmujoerfsmallL}{\tiny $-2.78 \pm 0.14$}
\newcommand{\mrmsendgpmujoerfsmallL}{\tiny $0.388 \pm 0.015$}

\newcommand{\lllvmAfbax}{\tiny $-34.93 \!\pm\! 10.12$}
\newcommand{\mrmselvmAfbax}{\tiny $0.583 \!\pm\! 0.015$}
\newcommand{\lllvmAkuka}{\tiny $-19.22 \!\pm\! 11.37$}
\newcommand{\mrmselvmAkuka}{\tiny $0.416 \!\pm\! 0.012$}
\newcommand{\lllvmAmujo}{\tiny $-40.09 \!\pm\! 4.16$}
\newcommand{\mrmselvmAmujo}{\tiny $0.748 \!\pm\! 0.009$}
\newcommand{\lllvmArbax}{\tiny $-18.29 \!\pm\! 2.75$}
\newcommand{\mrmselvmArbax}{\tiny $0.555 \!\pm\! 0.007$}
\newcommand{\lllvmAsarc}{\tiny $-32.58 \!\pm\! 6.48$}
\newcommand{\mrmselvmAsarc}{\tiny $0.654 \!\pm\! 0.011$}

\newcommand{\lllvmMxfbax}{\tiny $-49.69 \!\pm\! 0.21$}
\newcommand{\mrmselvmMxfbax}{\tiny $1.00 \!\pm\! 0.006$}
\newcommand{\lllvmMxkuka}{\tiny $-49.72 \!\pm\! 0.19$}
\newcommand{\mrmselvmMxkuka}{\tiny $1.001 \!\pm\! 0.006$}
\newcommand{\lllvmMxmujo}{\tiny $-43.28 \!\pm\! 6.44$}
\newcommand{\mrmselvmMxmujo}{\tiny $0.948 \!\pm\! 0.158$}
\newcommand{\lllvmMxrbax}{\tiny $-39.78 \!\pm\! 0.15$}
\newcommand{\mrmselvmMxrbax}{\tiny $1.001 \!\pm\! 0.005$}
\newcommand{\lllvmMxsarc}{\tiny $-39.76 \!\pm\! 0.10$}
\newcommand{\mrmselvmMxsarc}{\tiny $1.001 \!\pm\! 0.004$}

\newcommand{\lllvmMsAfbaxerft}{\tiny $-27.04 \!\pm\! 6.49$}
\newcommand{\mrmselvmMsAfbaxerft}{\tiny $0.522 \!\pm\! 0.038$}
\newcommand{\lllvmMsAfbaxerf}{\tiny $-25.87 \!\pm\! 4.43$}
\newcommand{\mrmselvmMsAfbaxerf}{\tiny $0.506 \!\pm\! 0.022$}
\newcommand{\lllvmMsAfbaxleakyrelu}{\tiny $-23.65 \!\pm\! 4.46$}
\newcommand{\mrmselvmMsAfbaxleakyrelu}{\tiny $0.483 \!\pm\! 0.028$}
\newcommand{\lllvmMsAfbaxrelu}{\tiny $-30.92 \!\pm\! 3.77$}
\newcommand{\mrmselvmMsAfbaxrelu}{\tiny $0.654 \!\pm\! 0.064$}
\newcommand{\lllvmMsAkukaerft}{\tiny $-23.59 \!\pm\! 12.63$}
\newcommand{\mrmselvmMsAkukaerft}{\tiny $0.388 \!\pm\! 0.024$}

\newcommand{\lllvmMsAkukaerf}{\tiny $-20.36 \!\pm\! 5.65$}
\newcommand{\mrmselvmMsAkukaerf}{\tiny $0.379 \!\pm\! 0.015$}
\newcommand{\lllvmMsAkukaleakyrelu}{\tiny $-17.00 \!\pm\! 9.57$}
\newcommand{\mrmselvmMsAkukaleakyrelu}{\tiny $0.373 \!\pm\! 0.027$}
\newcommand{\lllvmMsAkukarelu}{\tiny $-20.73 \!\pm\! 5.90$}
\newcommand{\mrmselvmMsAkukarelu}{\tiny $0.504 \!\pm\! 0.088$}
\newcommand{\lllvmMsAmujoerft}{\tiny $-42.71 \!\pm\! 6.23$}
\newcommand{\mrmselvmMsAmujoerft}{\tiny $0.768 \!\pm\! 0.033$}
\newcommand{\lllvmMsAmujoerf}{\tiny $-39.63 \!\pm\! 2.57$}
\newcommand{\mrmselvmMsAmujoerf}{\tiny $0.762 \!\pm\! 0.025$}

\newcommand{\lllvmMsAmujoleakyrelu}{\tiny $-35.40 \!\pm\! 2.26$}
\newcommand{\mrmselvmMsAmujoleakyrelu}{\tiny $0.691 \!\pm\! 0.015$}
\newcommand{\lllvmMsAmujorelu}{\tiny $-36.53 \!\pm\! 1.14$}
\newcommand{\mrmselvmMsAmujorelu}{\tiny $0.765 \!\pm\! 0.029$}
\newcommand{\lllvmMsArbaxerft}{\tiny $-13.98 \!\pm\! 3.14$}
\newcommand{\mrmselvmMsArbaxerft}{\tiny $0.500 \!\pm\! 0.035$}
\newcommand{\lllvmMsArbaxerf}{\tiny $-14.51 \!\pm\! 3.81$}
\newcommand{\mrmselvmMsArbaxerf}{\tiny $0.483 \!\pm\! 0.020$}
\newcommand{\lllvmMsArbaxleakyrelu}{\tiny $-13.45 \!\pm\! 1.78$}
\newcommand{\mrmselvmMsArbaxleakyrelu}{\tiny $0.489 \!\pm\! 0.021$}

\newcommand{\lllvmMsArbaxrelu}{\tiny $-30.60 \!\pm\! 3.41$}
\newcommand{\mrmselvmMsArbaxrelu}{\tiny $0.782 \!\pm\! 0.061$}
\newcommand{\lllvmMsAsarcerft}{\tiny $-28.51 \!\pm\! 3.76$}
\newcommand{\mrmselvmMsAsarcerft}{\tiny $0.633 \!\pm\! 0.039$}
\newcommand{\lllvmMsAsarcerf}{\tiny $-27.80 \!\pm\! 2.52$}
\newcommand{\mrmselvmMsAsarcerf}{\tiny $0.634 \!\pm\! 0.042$}
\newcommand{\lllvmMsAsarcleakyrelu}{\tiny $-26.75 \!\pm\! 5.33$}
\newcommand{\mrmselvmMsAsarcleakyrelu}{\tiny $0.569 \!\pm\! 0.030$}
\newcommand{\lllvmMsAsarcrelu}{\tiny $-30.06 \!\pm\! 2.05$}
\newcommand{\mrmselvmMsAsarcrelu}{\tiny $0.735 \!\pm\! 0.054$}

\newcommand{\lllvmMsMxfbaxerft}{\tiny $13.09 \!\pm\! 11.64$}
\newcommand{\mrmselvmMsMxfbaxerft}{\tiny $0.100 \!\pm\! 0.011$}
\newcommand{\lllvmMsMxfbaxerf}{\tiny $17.69 \!\pm\! 4.49$}
\newcommand{\mrmselvmMsMxfbaxerf}{\tiny $0.089 \!\pm\! 0.007$}
\newcommand{\lllvmMsMxfbaxleakyrelu}{\tiny $11.14 \!\pm\! 8.37$}
\newcommand{\mrmselvmMsMxfbaxleakyrelu}{\tiny $0.091 \!\pm\! 0.007$}
\newcommand{\lllvmMsMxfbaxrelu}{\tiny $19.49 \!\pm\! 6.64$}
\newcommand{\mrmselvmMsMxfbaxrelu}{\tiny $0.094 \!\pm\! 0.008$}
\newcommand{\lllvmMsMxkukaerft}{\tiny $15.60 \!\pm\! 5.30$}
\newcommand{\mrmselvmMsMxkukaerft}{\tiny $0.105 \!\pm\! 0.011$}

\newcommand{\lllvmMsMxkukaerf}{\tiny $22.18 \!\pm\! 2.07$}
\newcommand{\mrmselvmMsMxkukaerf}{\tiny $0.089 \!\pm\! 0.004$}
\newcommand{\lllvmMsMxkukaleakyrelu}{\tiny $15.65 \!\pm\! 6.65$}
\newcommand{\mrmselvmMsMxkukaleakyrelu}{\tiny $0.091 \!\pm\! 0.007$}
\newcommand{\lllvmMsMxkukarelu}{\tiny $22.82 \!\pm\! 1.89$}
\newcommand{\mrmselvmMsMxkukarelu}{\tiny $0.100 \!\pm\! 0.006$}
\newcommand{\lllvmMsMxmujoerft}{\tiny $-18.41 \!\pm\! 2.46$}
\newcommand{\mrmselvmMsMxmujoerft}{\tiny $0.405 \!\pm\! 0.016$}
\newcommand{\lllvmMsMxmujoerf}{\tiny $-19.10 \!\pm\! 2.22$}
\newcommand{\mrmselvmMsMxmujoerf}{\tiny $0.412 \!\pm\! 0.017$}

\newcommand{\lllvmMsMxmujoleakyrelu}{\tiny $-21.45 \!\pm\! 1.93$}
\newcommand{\mrmselvmMsMxmujoleakyrelu}{\tiny $0.422 \!\pm\! 0.013$}
\newcommand{\lllvmMsMxmujorelu}{\tiny $-18.06 \!\pm\! 0.57$}
\newcommand{\mrmselvmMsMxmujorelu}{\tiny $0.408 \!\pm\! 0.007$}
\newcommand{\lllvmMsMxrbaxerft}{\tiny $-1.16 \!\pm\! 3.95$}
\newcommand{\mrmselvmMsMxrbaxerft}{\tiny $0.353 \!\pm\! 0.007$}
\newcommand{\lllvmMsMxrbaxerf}{\tiny $1.88 \!\pm\! 1.36$}
\newcommand{\mrmselvmMsMxrbaxerf}{\tiny $0.348 \!\pm\! 0.004$}
\newcommand{\lllvmMsMxrbaxleakyrelu}{\tiny $1.33 \!\pm\! 2.13$}
\newcommand{\mrmselvmMsMxrbaxleakyrelu}{\tiny $0.347 \!\pm\! 0.005$}

\newcommand{\lllvmMsMxrbaxrelu}{\tiny $2.45 \!\pm\! 1.19$}
\newcommand{\mrmselvmMsMxrbaxrelu}{\tiny $0.355 \!\pm\! 0.005$}
\newcommand{\lllvmMsMxsarcerft}{\tiny $-9.63 \!\pm\! 5.32$}
\newcommand{\mrmselvmMsMxsarcerft}{\tiny $0.263 \!\pm\! 0.027$}
\newcommand{\lllvmMsMxsarcerf}{\tiny $-7.74 \!\pm\! 3.63$}
\newcommand{\mrmselvmMsMxsarcerf}{\tiny $0.259 \!\pm\! 0.021$}
\newcommand{\lllvmMsMxsarcleakyrelu}{\tiny $-9.44 \!\pm\! 3.26$}
\newcommand{\mrmselvmMsMxsarcleakyrelu}{\tiny $0.258 \!\pm\! 0.016$}
\newcommand{\lllvmMsMxsarcrelu}{\tiny $-5.50 \!\pm\! 1.81$}
\newcommand{\mrmselvmMsMxsarcrelu}{\tiny $0.257 \!\pm\! 0.014$}


\newcommand{\lllvmMsAfbaxerftdh}{\tiny $-23.20 \!\pm\! 9.07$}
\newcommand{\mrmselvmMsAfbaxerftdh}{\tiny $0.435 \!\pm\! 0.042$}
\newcommand{\lllvmMsAfbaxerfdh}{\tiny $-20.00 \!\pm\! 4.69$}
\newcommand{\mrmselvmMsAfbaxerfdh}{\tiny $0.428 \!\pm\! 0.028$}
\newcommand{\lllvmMsAfbaxleakyreludh}{\tiny $-18.88 \!\pm\! 5.10$}
\newcommand{\mrmselvmMsAfbaxleakyreludh}{\tiny $0.402 \!\pm\! 0.025$}
\newcommand{\lllvmMsAfbaxreludh}{\tiny $-28.87 \!\pm\! 5.06$}
\newcommand{\mrmselvmMsAfbaxreludh}{\tiny $0.602 \!\pm\! 0.082$}
\newcommand{\lllvmMsAkukaerftdh}{\tiny $-28.77 \!\pm\! 16.68$}
\newcommand{\mrmselvmMsAkukaerftdh}{\tiny $0.363 \!\pm\! 0.028$}
\newcommand{\lllvmMsAkukaerfdh}{\tiny $-23.47 \!\pm\! 10.88$}
\newcommand{\mrmselvmMsAkukaerfdh}{\tiny $0.349 \!\pm\! 0.024$}
\newcommand{\lllvmMsAkukaleakyreludh}{\tiny $-11.52 \!\pm\! 5.78$}
\newcommand{\mrmselvmMsAkukaleakyreludh}{\tiny $0.301 \!\pm\! 0.017$}
\newcommand{\lllvmMsAkukareludh}{\tiny $-18.33 \!\pm\! 5.13$}
\newcommand{\mrmselvmMsAkukareludh}{\tiny $0.479 \!\pm\! 0.086$}
\newcommand{\lllvmMsAmujoerftdh}{\tiny $-35.62 \!\pm\! 1.99$}
\newcommand{\mrmselvmMsAmujoerftdh}{\tiny $0.679 \!\pm\! 0.035$}
\newcommand{\lllvmMsAmujoerfdh}{\tiny $-34.69 \!\pm\! 1.96$}
\newcommand{\mrmselvmMsAmujoerfdh}{\tiny $0.679 \!\pm\! 0.037$}
\newcommand{\lllvmMsAmujoleakyreludh}{\tiny $-32.94 \!\pm\! 2.05$}
\newcommand{\mrmselvmMsAmujoleakyreludh}{\tiny $0.626 \!\pm\! 0.018$}
\newcommand{\lllvmMsAmujoreludh}{\tiny $-32.32 \!\pm\! 0.88$}
\newcommand{\mrmselvmMsAmujoreludh}{\tiny $0.670 \!\pm\! 0.021$}
\newcommand{\lllvmMsArbaxerftdh}{\tiny $-15.87 \!\pm\! 6.49$}
\newcommand{\mrmselvmMsArbaxerftdh}{\tiny $0.461 \!\pm\! 0.016$}
\newcommand{\lllvmMsArbaxerfdh}{\tiny $-18.65 \!\pm\! 7.67$}
\newcommand{\mrmselvmMsArbaxerfdh}{\tiny $0.490 \!\pm\! 0.030$}
\newcommand{\lllvmMsArbaxleakyreludh}{\tiny $-12.72 \!\pm\! 2.45$}
\newcommand{\mrmselvmMsArbaxleakyreludh}{\tiny $0.461 \!\pm\! 0.011$}
\newcommand{\lllvmMsArbaxreludh}{\tiny $-26.91 \!\pm\! 3.42$}
\newcommand{\mrmselvmMsArbaxreludh}{\tiny $0.726 \!\pm\! 0.064$}
\newcommand{\lllvmMsAsarcerftdh}{\tiny $-23.58 \!\pm\! 2.81$}
\newcommand{\mrmselvmMsAsarcerftdh}{\tiny $0.520 \!\pm\! 0.024$}
\newcommand{\lllvmMsAsarcerfdh}{\tiny $-24.08 \!\pm\! 2.62$}
\newcommand{\mrmselvmMsAsarcerfdh}{\tiny $0.536 \!\pm\! 0.034$}
\newcommand{\lllvmMsAsarcleakyreludh}{\tiny $-24.43 \!\pm\! 2.17$}
\newcommand{\mrmselvmMsAsarcleakyreludh}{\tiny $0.500 \!\pm\! 0.018$}
\newcommand{\lllvmMsAsarcreludh}{\tiny $-27.36 \!\pm\! 2.23$}
\newcommand{\mrmselvmMsAsarcreludh}{\tiny $0.652 \!\pm\! 0.038$}


\newcommand{\lllvmMsAfbaxerftth}{\tiny $-24.61 \!\pm\! 10.86$}
\newcommand{\mrmselvmMsAfbaxerftth}{\tiny $0.415 \!\pm\! 0.035$}
\newcommand{\lllvmMsAfbaxerfth}{\tiny $-22.06 \!\pm\! 3.42$}
\newcommand{\mrmselvmMsAfbaxerfth}{\tiny $0.437 \!\pm\! 0.030$}
\newcommand{\lllvmMsAfbaxleakyreluth}{\tiny $-23.97 \!\pm\! 7.54$}
\newcommand{\mrmselvmMsAfbaxleakyreluth}{\tiny $0.384 \!\pm\! 0.030$}
\newcommand{\lllvmMsAfbaxreluth}{\tiny $-25.80 \!\pm\! 4.62$}
\newcommand{\mrmselvmMsAfbaxreluth}{\tiny $0.551 \!\pm\! 0.077$}
\newcommand{\lllvmMsAkukaerftth}{\tiny $-29.75 \!\pm\! 14.93$}
\newcommand{\mrmselvmMsAkukaerftth}{\tiny $0.343 \!\pm\! 0.041$}
\newcommand{\lllvmMsAkukaerfth}{\tiny $-23.10 \!\pm\! 12.28$}
\newcommand{\mrmselvmMsAkukaerfth}{\tiny $0.344 \!\pm\! 0.018$}
\newcommand{\lllvmMsAkukaleakyreluth}{\tiny $-13.27 \!\pm\! 9.36$}
\newcommand{\mrmselvmMsAkukaleakyreluth}{\tiny $0.294 \!\pm\! 0.024$}
\newcommand{\lllvmMsAkukareluth}{\tiny $-16.14 \!\pm\! 5.31$}
\newcommand{\mrmselvmMsAkukareluth}{\tiny $0.450 \!\pm\! 0.082$}
\newcommand{\lllvmMsAmujoerftth}{\tiny $-33.60 \!\pm\! 2.60$}
\newcommand{\mrmselvmMsAmujoerftth}{\tiny $0.637 \!\pm\! 0.029$}
\newcommand{\lllvmMsAmujoerfth}{\tiny $-34.68 \!\pm\! 4.49$}
\newcommand{\mrmselvmMsAmujoerfth}{\tiny $0.643 \!\pm\! 0.034$}
\newcommand{\lllvmMsAmujoleakyreluth}{\tiny $-35.15 \!\pm\! 3.19$}
\newcommand{\mrmselvmMsAmujoleakyreluth}{\tiny $0.628 \!\pm\! 0.027$}
\newcommand{\lllvmMsAmujoreluth}{\tiny $-32.07 \!\pm\! 1.25$}
\newcommand{\mrmselvmMsAmujoreluth}{\tiny $0.659 \!\pm\! 0.023$}
\newcommand{\lllvmMsArbaxerftth}{\tiny $-29.10 \!\pm\! 8.81$}
\newcommand{\mrmselvmMsArbaxerftth}{\tiny $0.457 \!\pm\! 0.021$}
\newcommand{\lllvmMsArbaxerfth}{\tiny $-12.34 \!\pm\! 3.75$}
\newcommand{\mrmselvmMsArbaxerfth}{\tiny $0.468 \!\pm\! 0.028$}
\newcommand{\lllvmMsArbaxleakyreluth}{\tiny $-12.85 \!\pm\! 7.28$}
\newcommand{\mrmselvmMsArbaxleakyreluth}{\tiny $0.457 \!\pm\! 0.022$}
\newcommand{\lllvmMsArbaxreluth}{\tiny $-23.95 \!\pm\! 3.86$}
\newcommand{\mrmselvmMsArbaxreluth}{\tiny $0.668 \!\pm\! 0.078$}
\newcommand{\lllvmMsAsarcerftth}{\tiny $-24.46 \!\pm\! 3.51$}
\newcommand{\mrmselvmMsAsarcerftth}{\tiny $0.500 \!\pm\! 0.023$}
\newcommand{\lllvmMsAsarcerfth}{\tiny $-22.32 \!\pm\! 2.33$}
\newcommand{\mrmselvmMsAsarcerfth}{\tiny $0.502 \!\pm\! 0.028$}
\newcommand{\lllvmMsAsarcleakyreluth}{\tiny $-30.14 \!\pm\! 10.75$}
\newcommand{\mrmselvmMsAsarcleakyreluth}{\tiny $0.520 \!\pm\! 0.061$}
\newcommand{\lllvmMsAsarcreluth}{\tiny $-24.32 \!\pm\! 1.66$}
\newcommand{\mrmselvmMsAsarcreluth}{\tiny $0.600 \!\pm\! 0.044$}

In this section we conduct a series of experiments to illustrate the modeling potential of the models described in Sec.~\ref{sec:models}.
First, in Sec.~\ref{sec:synthetic} we conduct a simple experiment with synthetic data.
Next, in Sec.~\ref{sec:data} we describe the robotics data that we use in all our remaining experiments.
In Sec.~\ref{sec:regexp} we consider regression models, while in Sec.~\ref{sec:lvmexp} we consider the unsupervised setting.
In addition in Sec.~\ref{sec:varynh} we consider the effect of varying the number of `hidden units' in the neural likelihood, while in
Sec.~\ref{sec:smalldata}-\ref{sec:missing} we examine the small data regime and missing outputs, respectively.
All our experiments are implemented using GPyTorch \citep{gardner2018gpytorch} and PyTorch \citep{paszke2017automatic}.

\subsection{Synthetic Regression Experiment}
\label{sec:synthetic}
\begin{figure}[t!]
    \centering
    \protect\begin{subfigure}{.50\textwidth}
        \includegraphics[width=\textwidth]{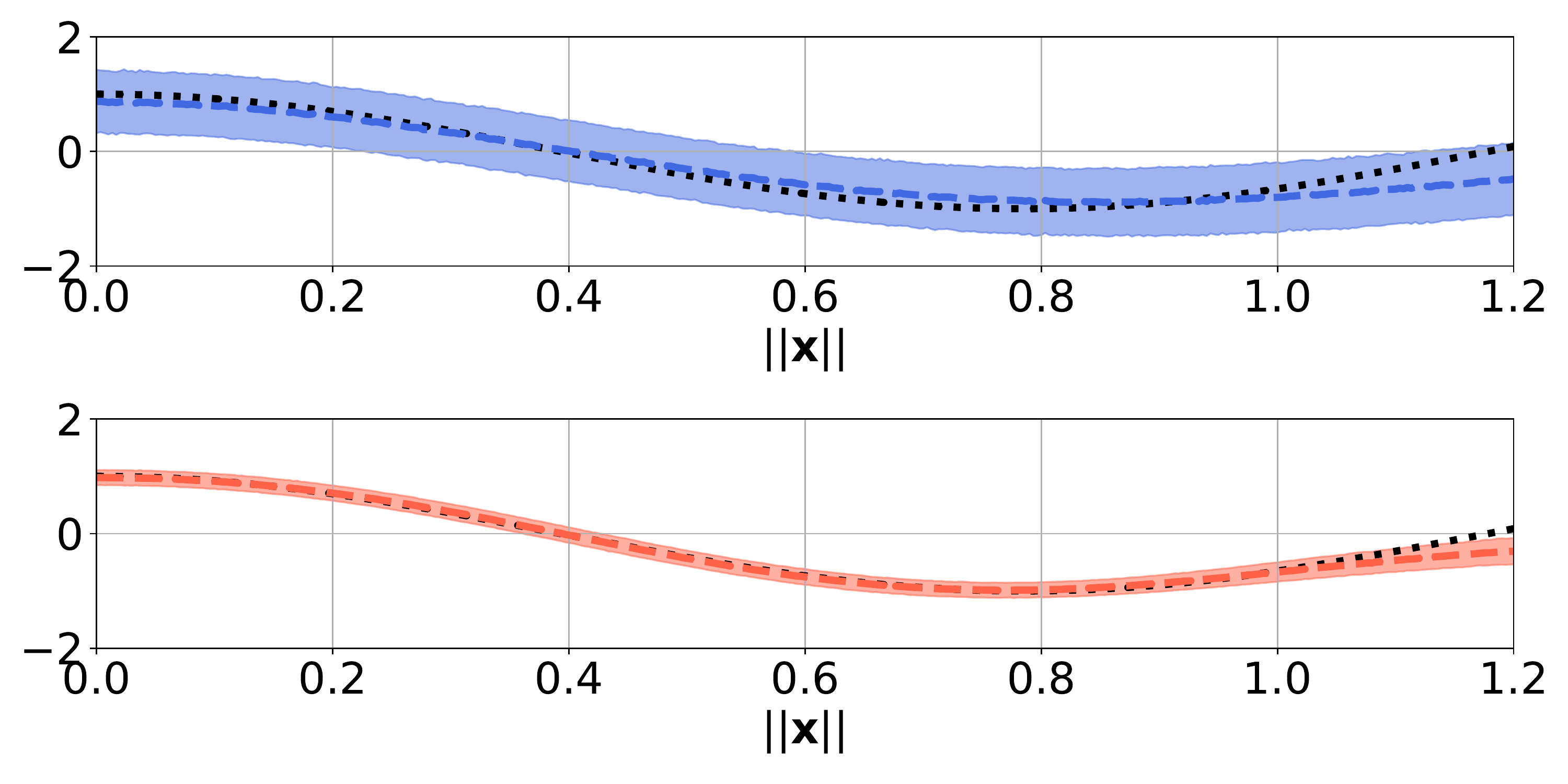}
    \protect\end{subfigure}\hfill
    \caption{Predictions for the MOGP (top) and N-MOGP (bottom) for the synthetic regression experiment in Sec.~\ref{sec:synthetic}.
    We depict the true function with a black dotted line, the mean model predictions with dashed lines, and colored uncertainty bands that extend from the $10^{\rm th}$ to the $90^{\rm th}$ percentile. Note that predictions for $||\bx||>1$ are extrapolations.}
    \label{fig:synthetic}
\end{figure}

We conduct a simple experiment using synthetic data to explore the modeling capacity of neural likelihoods. We consider the function
$g : \RR^5 \to \RR^8$ given by $g_k(\bx) = \cos(4 || \bx||)$ for $k=1,...,8$ where $ || \bx||$ is the L2-norm of $\bx$. We sample
$N=1000$ inputs $\{\bx_i\}$ from the unit ball in $\RR^5$ and generate a dataset with noisy outputs via
$\mathcal{D} = \{(\bx_i, g(\bx_i) + \sigma_0 \beps_i) \}$ with $\beps_i \sim \mathcal{N}({\bf 0}, {\bf 1})$ and where $\sigma_0=0.1$.
We then compare the quality of fit obtained by a MOGP versus a Neural MOGP. For both models we set the number of GPs to $L=3$, use
$N_{\rm ind}=200$ inducing points, and choose
$D_H=8$ for the N-MOGP.

To assess the quality of the fit visually, we choose a random line segment in $\RR^5$ originating at the
origin as well as a random output dimension $k \in [1, 8]$ and
depict model predictions $y_k(\bx^*)$ along the line segment, see Fig.~\ref{fig:synthetic}. While both models are able
to learn reasonable mean functions, the MOGP exhibits a higher test MRMSE (0.166) than the N-MOGP (0.102). More strikingly,
the N-MOGP is able to learn better calibrated uncertainties and thus obtains a substantially
higher test log likelihood: 6.92 versus -0.72.
One reason for this difference may be due to our choice of a bounded\footnote{Specifically we chose
the (shifted) error function $\sigma(x) = {\rm erf}(x) + 1$.} non-linearity $\sigma(\cdot)$, which gives the N-MOGP more flexibility in learning
a suitable variance function.\footnote{Recall from Eqn.~\ref{eqn:regressorell} that in order for a regression model to obtain a large expected log likelihood it
must learn a high-quality mean function \emph{and} a high-quality variance function.}


\subsection{Data}
\label{sec:data}
We use five robotics datasets for our main set of  experiments, four of which were collected from real-world robots and one of which was generated
using the MuJoCo physics simulator \citep{todorov2012mujoco}. These datasets have been used in a number of papers,
including references \citep{vijayakumar2000locally,meier2014incremental,cheng2017variational}. In all five datasets the input and output dimensions correspond to various joint positions/velocities/etc.~of the robot. These datasets form a good testbed for our proposed models, since the complex dynamics recorded in these data is highly non-linear and inherently multi-dimensional. See Table~\ref{table:data} for a summary of the different datasets.

\begin{table}[h]
\begin{center}
\resizebox{.32 \textwidth}{!}{
    \begin{tabu}{|c|[1pt]c|c|c|c|}    \hline
    \small Dataset \cellcolor[gray]{0.75} & \small $N_{\rm train}$ \cellcolor[gray]{0.95}& \small $N_{\rm test}$ \cellcolor[gray]{0.95} & \small $D_X$ \cellcolor[gray]{0.95}& \small $D_Y$ \cellcolor[gray]{0.95}\\  \tabucline[1pt]{-}
    \small R-Baxter \cellcolor[gray]{0.95}&\small  6918& \small  2000 & \small 21 & \small 7\\ \hline
    \small F-Baxter \cellcolor[gray]{0.95}&\small  14295 & \small  5000 & \small 21& \small14 \\ \hline
    \small Kuka \cellcolor[gray]{0.95}&\small 15068 & \small  5000 & \small 21& \small 14\\ \hline
    \small Sarcos \cellcolor[gray]{0.95}&\small  43933 & \small 5000 & \small 21& \small 7\\ \hline
    \small MuJoCo \cellcolor[gray]{0.95}&\small  $10^5$ & \small $10^4$ & \small 23& \small9 \\ \hline
    \end{tabu}
    } 
\end{center}
     \caption{Datasets used in our experiments.} 
     \label{table:data}
\end{table}

\begin{table*}[t]
\centering
\tabcolsep=0.11cm
\resizebox{1 \textwidth}{!}{
\begin{tabular}{|*{11}{c|}}
\hline
 \cellcolor[gray]{0.65}& \multicolumn{10}{c|}{\cellcolor[gray]{0.80}{\bf Dataset}} \\
   \cellcolor[gray]{0.65}& \multicolumn{2}{c|}{{\bf R-Baxter}\cellcolor[gray]{0.95}} & \multicolumn{2}{c|}{{\bf F-Baxter}\cellcolor[gray]{0.95}} & \multicolumn{2}{c|}{{\bf Kuka}\cellcolor[gray]{0.95}} & \multicolumn{2}{c|}{{\bf Sarcos}\cellcolor[gray]{0.95}} & \multicolumn{2}{c|}{{\bf MuJoCo}\cellcolor[gray]{0.95}}\\
  \Xcline{1-11}{0.5pt}
 \small {\bf Model } \cellcolor[gray]{0.80} & \tiny LL & \tiny MRMSE & \tiny LL & \tiny MRMSE & \tiny LL & \tiny MRMSE & \tiny LL & \tiny MRMSE & \tiny LL & \tiny MRMSE\\ 
\Xcline{1-11}{0.5pt}\hline\hline
 \multicolumn{11}{|c|}{\tiny {\bf Baseline Models}\cellcolor[gray]{0.90}} \\ \hline\hline
\lgc  \tiny MOGP &  \llArbax & \mrmseArbax & \llAfbax & \mrmseAfbax & \llAkuka & \mrmseAkuka & \llAsarc & \mrmseAsarc & \llAmujo & \mrmseAmujo \\ \hline
\lgc  \tiny DK-MOGP &  \llArbaxdk & \mrmseArbaxdk & \llAfbaxdk & \mrmseAfbaxdk & \llAkukadk & \mrmseAkukadk & \llAsarcdk & \mrmseAsarcdk & \llAmujodk & \mrmseAmujodk  \\ \hline  \hline
 \lgc  \tiny GPRN &  \llgprnrbax & \mrmsegprnrbax & \llgprnfbax & \mrmsegprnfbax & \llgprnkuka & \mrmsegprnkuka & \llgprnsarc & \mrmsegprnsarc & \llgprnmujo & \mrmsegprnmujo \\ \hline
  \lgc  \tiny DK-GPRN &  \llgprnrbaxdk & \mrmsegprnrbaxdk & \llgprnfbaxdk & \mrmsegprnfbaxdk & \llgprnkukadk & \mrmsegprnkukadk & \llgprnsarcdk & \mrmsegprnsarcdk & \llgprnmujodk & \mrmsegprnmujodk \\ \hline \hline
  \lgc  \tiny DGP &  \lldeepgprbax & \mrmsedeepgprbax & \lldeepgpfbax & \mrmsedeepgpfbax & \lldeepgpkuka & \mrmsedeepgpkuka & \lldeepgpsarc & \mrmsedeepgpsarc & \lldeepgpmujo & \mrmsedeepgpmujo \\ \hline \hline
 \multicolumn{11}{|c|}{\tiny {\bf Neural Likelihood Models}\cellcolor[gray]{0.90}} \\ \hline\hline
\lgc    \tiny N-MOGP&  \llMsArbaxerftdh & \mrmseMsArbaxerftdh & \llMsAfbaxerftdh & \mrmseMsAfbaxerftdh & \llMsAkukaerftdh & \mrmseMsAkukaerftdh & \llMsAsarcerftdh & \mrmseMsAsarcerftdh & \llMsAmujoerftdh & \mrmseMsAmujoerftdh \\ \hline \hline
\lgc  \tiny SBGPRN &  \llMxrbax & \mrmseMxrbax & \llMxfbax & \mrmseMxfbax & \llMxkuka & \mrmseMxkuka & \llMxsarc & \mrmseMxsarc & \llMxmujo & \mrmseMxmujo \\ \hline
\lgc   \tiny N-SBGPRN&  \llMsMxrbaxleakyrelu & \mrmseMsMxrbaxleakyrelu & \llMsMxfbaxleakyrelu & \mrmseMsMxfbaxleakyrelu & \llMsMxkukaleakyrelu & \mrmseMsMxkukaleakyrelu & \llMsMxsarcleakyrelu & \mrmseMsMxsarcleakyrelu & \llMsMxmujoleakyrelu & \mrmseMsMxmujoleakyrelu \\ \hline
    \lgc \tiny DK-N-SBGPRN &  \llMsMxrbaxleakyreludk & \mrmseMsMxrbaxleakyreludk & \llMsMxfbaxleakyreludk & \mrmseMsMxfbaxleakyreludk & \llMsMxkukaleakyreludk & \mrmseMsMxkukaleakyreludk & \llMsMxsarcleakyreludk & \mrmseMsMxsarcleakyreludk & \llMsMxmujoleakyreludk & \mrmseMsMxmujoleakyreludk \\ \hline \hline
  \lgc  \tiny N-DGP &  \llndgprbaxerfdh & \mrmsendgprbaxerfdh & \llndgpfbaxerfdh & \mrmsendgpfbaxerfdh & \llndgpkukaerfdh & \mrmsendgpkukaerfdh & \llndgpsarcerfdh & \mrmsendgpsarcerfdh & \llndgpmujoerfdh & \mrmsendgpmujoerfdh \\ \hline 
\end{tabular}
} 
\caption{Results for the regression experiments in Sec.~\ref{sec:regexp}. We report test log likelihoods per datapoint (LL) and mean root mean squared errors\protect\footnotemark (MRMSE) averaged over ten random train/test splits of the data. See the supplementary materials for additional results and model details.}.
  \label{table:reg}
\end{table*}
\footnotetext{MRMSE is the RMSE along each output dimension averaged across all output dimensions.}

\subsection{Regression}
\label{sec:regexp}

In this section we compare the performance of the various regression models defined in Sec.~\ref{sec:regmodels}.
To facilitate a fair comparison we choose the same number of GPs $L$ in all models.
In particular we choose $L=\ceil{D_Y/2}$, where $\ceil{\cdot}$ denotes the ceiling function.
For the  (N-)DGP models we choose $L^\prime = \ceil{\tfrac{3}{4}D_Y}$ so that the DGP prior is expected to be quite flexible.
For the N-MOGP, N-SBGPRN, and N-DGP models, each of which includes a hyperparameter controlling the number of hidden units,
we set $D_H = 2D_Y$, $D_H = D_Y$, and $D_H =2D_Y$, respectively.
For several of the models we also report results with a deep kernel (denoted by `DK').
For the models with neural likelihoods, we experiment with the following set of non-linearities $\sigma(\cdot)$:
\begin{enumerate}
\item ReLU: ${\rm relu}(x) \equiv \max(0, x)$
\item Leaky ReLU: ${\rm leaky}(x) \equiv \max(\epsilon x, x)$ with\footnote{We choose $\epsilon=0.35$ in our experiments.} $\epsilon>0$
\item Error function: ${\rm erf}(x)$
\item Shifted error function: ${\rm sherf}(x) \equiv 1 + {\rm erf}(x)$
\end{enumerate}
 For a partial set of results see Table \ref{table:reg}, which is organized to facilitate comparison between baseline models and their neural counterparts (e.g.~MOGP versus N-MOGP). For additional details on the models and for additional results see the supplementary materials.

For most of the models and datasets predictive performance improves substantially with the addition of a neural likelihood; this is especially pronounced for the N-MOGP and (N-)SBGPRN. For the flexible DGP prior the gain in performance tends to be smaller (although see the R-Baxter and F-Baxter datasets). This smaller gain in performance, however, is largely a result of our choice of $L^\prime$. Indeed if we choose $L=L^\prime$ (so that the DGP prior is less powerful) the performance jump from DGP to N-DGP is substantial;
see the supplementary materials.
 Note that in one case (F-Baxter) the N-MOGP has better predictive performance than the DGP and in three out of five datasets N-MOGP outperforms the GPRN in log likelihood, even though the GPRN achieves
 higher LLs than the MOGP on all five datasets. The N-SBGPRN and DK-N-SBGPRN perform particularly well across all five
 datasets; this is encouraging because we found these models easy and fast to train. Note as well that with the DK-N-SBGPRN
 we demonstrate that neural likelihoods can be successfuly combined with deep kernels.

\subsection{Unsupervised Learning}
\label{sec:lvmexp}

\begin{table*}[t!]
\centering
\tabcolsep=0.11cm
\resizebox{1 \textwidth}{!}{
\begin{tabular}{|*{11}{c|}}
\hline
 \cellcolor[gray]{0.65}& \multicolumn{10}{c|}{\cellcolor[gray]{0.80}{\bf Dataset}} \\
   \cellcolor[gray]{0.65}& \multicolumn{2}{c|}{{\bf R-Baxter}\cellcolor[gray]{0.95}} & \multicolumn{2}{c|}{{\bf F-Baxter}\cellcolor[gray]{0.95}} & \multicolumn{2}{c|}{{\bf Kuka}\cellcolor[gray]{0.95}} & \multicolumn{2}{c|}{{\bf Sarcos}\cellcolor[gray]{0.95}} & \multicolumn{2}{c|}{{\bf MuJoCo}\cellcolor[gray]{0.95}}\\
  \Xcline{1-11}{0.5pt}
 \small {\bf Model}  \cellcolor[gray]{0.80} & \tiny LL & \tiny MRMSE & \tiny LL & \tiny MRMSE & \tiny LL & \tiny MRMSE & \tiny LL & \tiny MRMSE & \tiny LL & \tiny MRMSE\\ 
\Xcline{1-11}{0.5pt}
\lgc  \tiny MOGP &  \lllvmArbax & \mrmselvmArbax & \lllvmAfbax & \mrmselvmAfbax & \lllvmAkuka & \mrmselvmAkuka & \lllvmAsarc & \mrmselvmAsarc & \lllvmAmujo & \mrmselvmAmujo \\ \hline
\lgc   \tiny N-MOGP&  \lllvmMsArbaxleakyreludh & \mrmselvmMsArbaxleakyreludh & \lllvmMsAfbaxleakyreludh & \mrmselvmMsAfbaxleakyreludh & \lllvmMsAkukaleakyreludh & \mrmselvmMsAkukaleakyreludh & \lllvmMsAsarcleakyreludh & \mrmselvmMsAsarcleakyreludh & \lllvmMsAmujoleakyreludh & \mrmselvmMsAmujoleakyreludh \\\hline
\lgc   \tiny N-SBGPRN&  \lllvmMsMxrbaxrelu & \mrmselvmMsMxrbaxrelu & \lllvmMsMxfbaxrelu & \mrmselvmMsMxfbaxrelu & \lllvmMsMxkukarelu & \mrmselvmMsMxkukarelu & \lllvmMsMxsarcrelu & \mrmselvmMsMxsarcrelu & \lllvmMsMxmujorelu & \mrmselvmMsMxmujorelu \\ \hline
\end{tabular}
} 
\caption{Results for the unsupervised learning experiments in Sec.~\ref{sec:lvmexp}. We report test log likelihoods per datapoint (LL) and mean root mean squared errors (MRMSE) averaged over ten random train/test splits of the data.
See the supplementary materials for additional results and model details.}.
  \label{table:lvm}
\end{table*}

In this section we compare the performance of several unsupervised models defined by the general recipe in Sec.~\ref{sec:lvmmodels}.
In particular we consider unsupervised versions of the following models:  MOGP,  N-MOGP, and  N-SBGPRN. We
also trained unsupervised versions of the GPRN, DPG, and SBGPRN, but we do not report any results, since we found these models to perform poorly.\footnote{In order to get a deep GP with latent inputs to achieve good performance we would presumably need to implement a custom inference procedure more along the lines of the one used in \citep{dai2015variational}. The sampling-based approach we used struggled to learn anything, probably at least in part due to high variance gradients.} Note that we turn the supervised datasets described in Sec.~\ref{sec:data}
into unsupervised datasets by concatenating the inputs and outputs: $\by_i \leftarrow (\bx_i, \by_i)$.

For all models we set the latent dimension to $D_X=4$ and the number of GPs to $L=4$. For the N-MOGP and N-SBGPRN we set $D_H=14$ and $D_H=7$, respectively.
For a partial set of results see Table \ref{table:lvm}.
For additional details on the models and for additional results see the supplementary materials.

Analogous to the regression models in the previous section, we find that the models with neural likelihoods substantially outperform the baseline MOGP. The performance gain is especially striking for the N-SBGPRN,
which is the clear winner on all five datasets. This result is somewhat surprising, in that one might worry that the N-SBGPRN---which employs a deterministic neural network to mix the latent Gaussian processes in the
likelihood---could be especially susceptible to overfitting in the unsupervised setting. In fact, while we do see evidence\footnote{We typically find a difference of about 1 nat between train and test log likelihoods (here normalized per output dimension).} of moderate overfitting on these datasets, we find that the increased flexibility of the likelihood easily compensates for any loss in performance caused by overfitting. This result is encouraging because (as above) we generally found the N-SBGPRN easy and fast to train.

\subsection{Varying the Number of Hidden Units}
\label{sec:varynh}

\begin{figure}[h!]
    \centering
    \begin{subfigure}{.49\textwidth}
        \includegraphics[width=.85\textwidth]{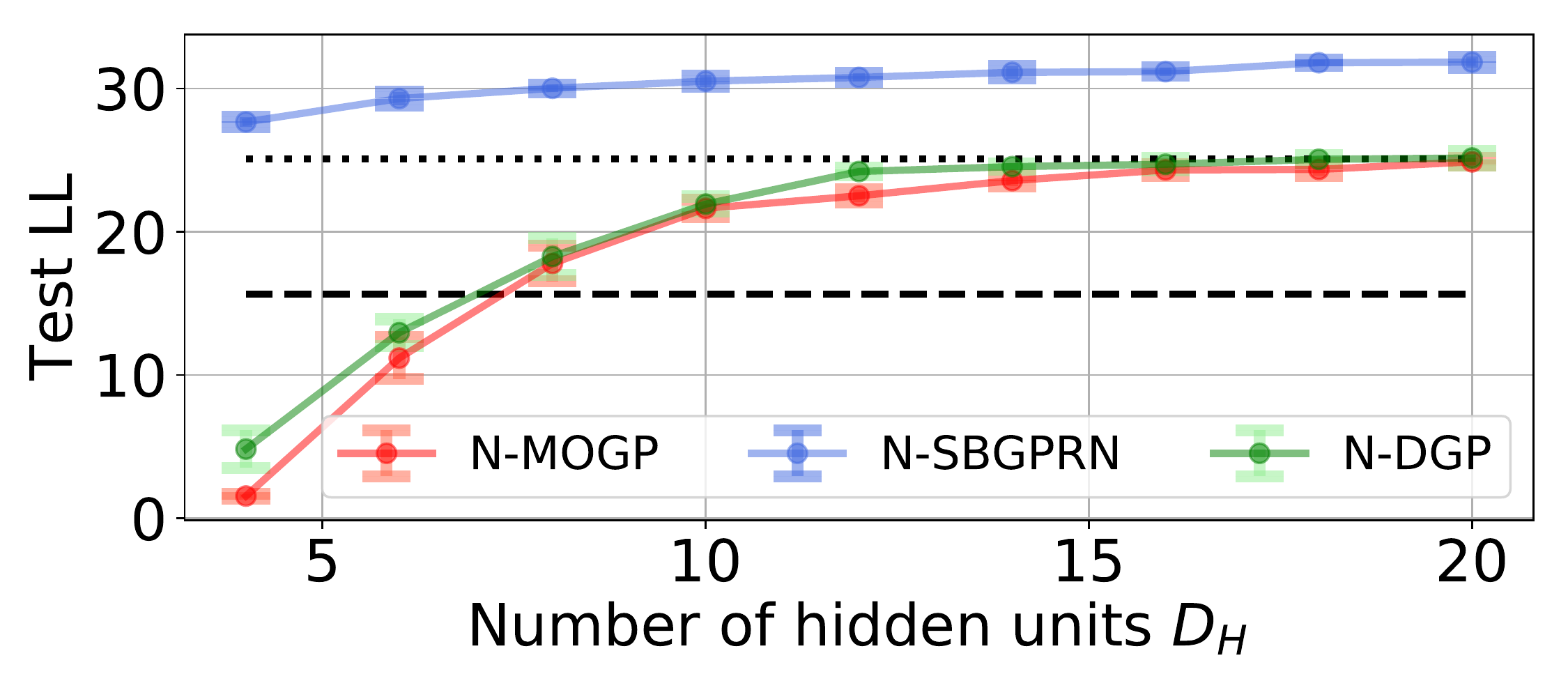}
    \end{subfigure}\hfill
    \begin{subfigure}{.49\textwidth}
        \includegraphics[width=.85\textwidth]{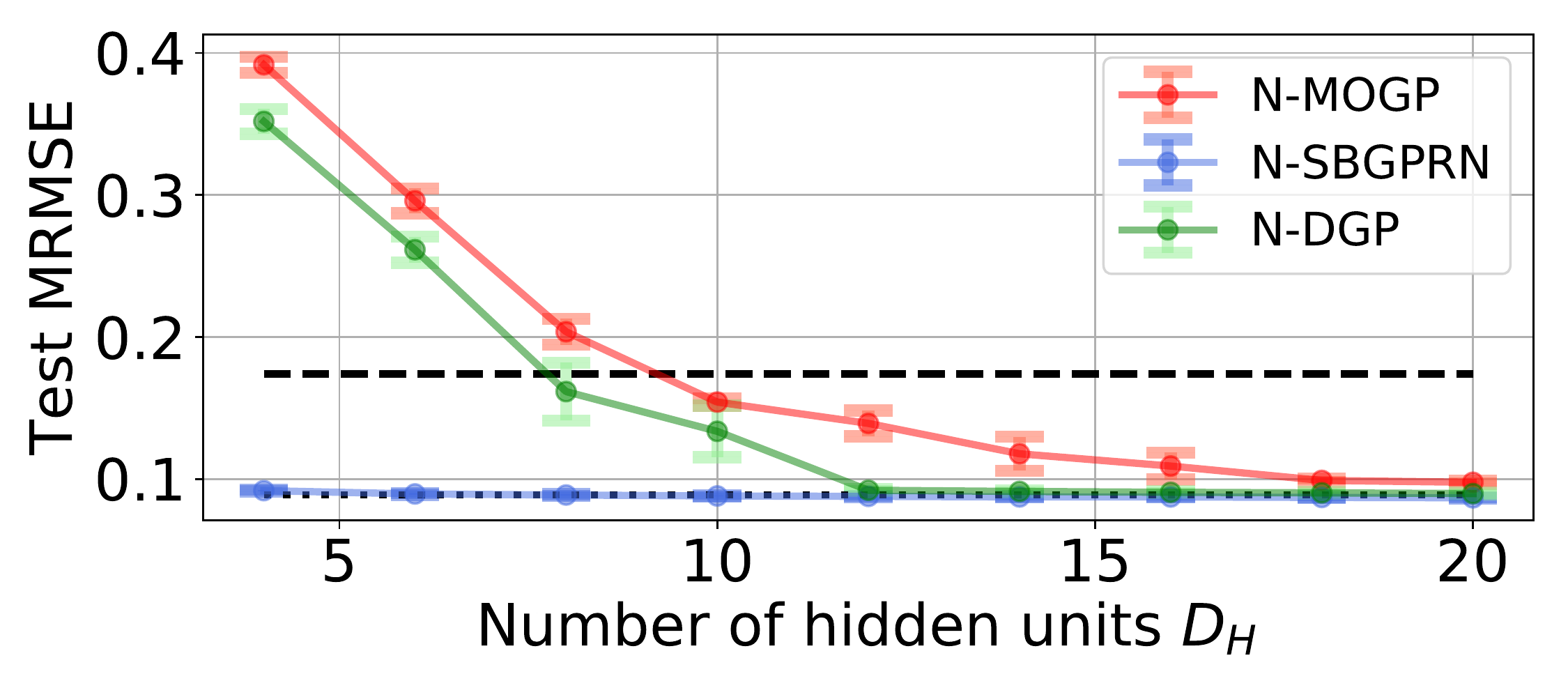}
    \end{subfigure}
    \begin{subfigure}{.49\textwidth}
        \includegraphics[width=.85\textwidth]{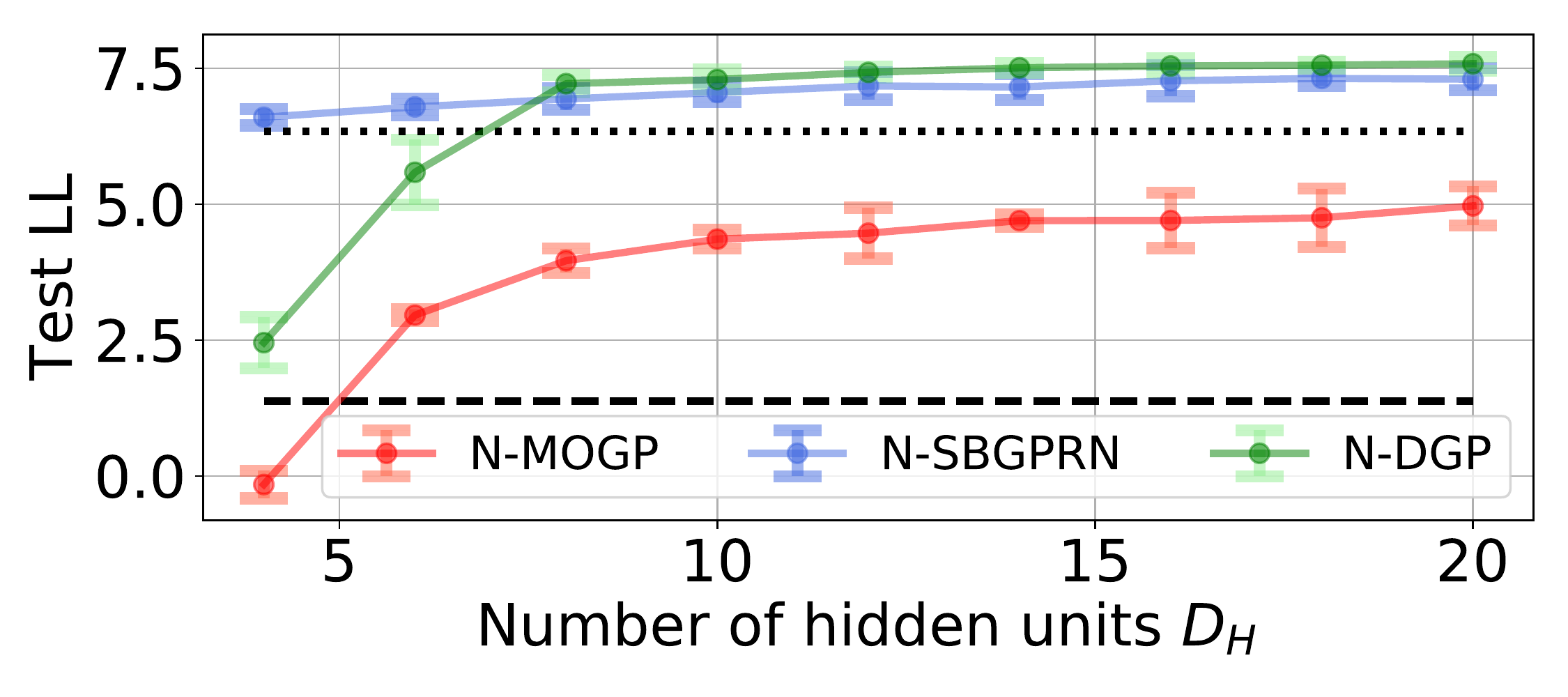}
    \end{subfigure}\hfill
    \begin{subfigure}{.49\textwidth}
        \includegraphics[width=.85\textwidth]{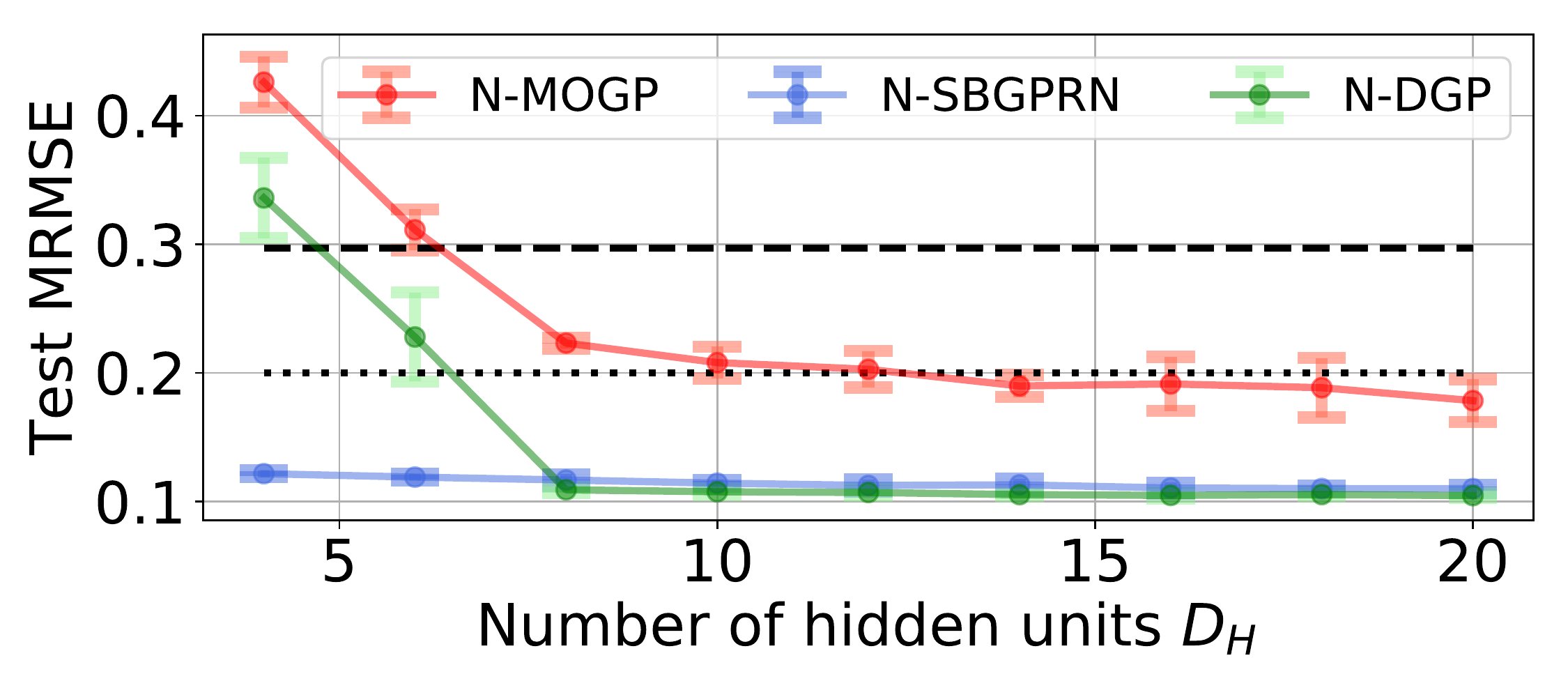}
    \end{subfigure}
    \caption{Test LLs and MRMSEs as a function of the number of hidden units on the  Kuka (top) and R-Baxter (bottom) datasets for three neural GP models. For comparison we include results for the MOGP (dashed line) and DGP (dotted line). Results are averaged over ten random train/test splits.}
    \label{fig:varynh}
\end{figure}

To explore the effect of varying the number of hidden units $D_H$ we train
N-MOGP, N-SBGPRN, and N-DGP regression models on the Kuka and R-Baxter datasets for a range of
$D_H \in [4, 20]$. See Fig.~\ref{fig:varynh} for the results. As we would expect,
we find that the performance---both in terms of the test log likelihood and the test
MRMSE---tends to improve for all three models as we increase the number of hidden units.
However, the effect is much more pronounced for the N-MOGP and N-DGP, where the likelihoods are
not as flexible as in the N-SBGPRN, which includes a (deterministic) neural network as a
subcomponent. Notably, as we increase the number of hidden units in the N-MOGP and N-DGP
we close the majority or all of the performance gap between these two models and the N-SBGPRN.
This is encouraging, since we expect the
N-MOGP and N-DGP to be less prone to overfitting.

\subsection{Small Data Regime}
\label{sec:smalldata}

\begin{figure}[t!]
    \centering
    \begin{subfigure}{.49\textwidth}
        \includegraphics[width=.99\textwidth]{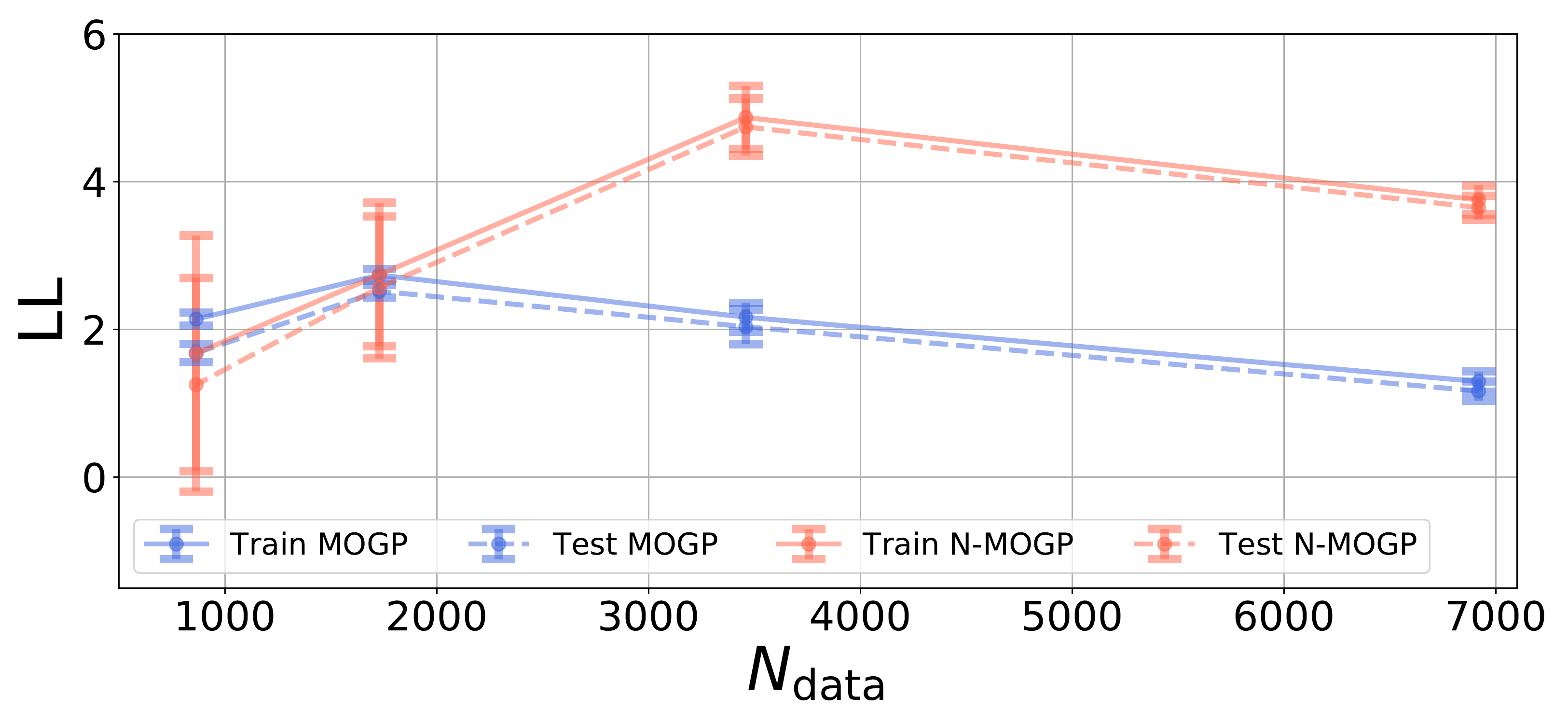}
    \end{subfigure}\hfill
    \caption{Training and Test LLs for models trained on varying amounts of training data $N_{\rm data}$ for the R-Baxter dataset. Results are averaged over fifteen random train/test splits for each value of $N_{\rm data}$.}
    \label{fig:smalldata}
\end{figure}

Here we explore the extent to which the models defined in Sec.~\ref{sec:regmodels} are susceptible to overfitting.
Among the models with neural likelihoods, we choose the N-MOGP, since, as discussed above, we expect it to be
robust in the small data regime.\footnote{In addition we find that the SBGPRN  and (N-)SBGPRN are actually susceptible to \emph{underfitting} in this regime because of a tendency to get stuck in bad local minima.}
We then compare the N-MOGP to the MOGP and depict train and test log likelihoods obtained on the R-Baxter dataset as we vary the amount of training data, see Fig.~\ref{fig:smalldata}. Although, as expected, we tend to observe lower log likelihoods as the number of training datapoints decreases, there is no evidence for overfitting.
We observe similar results for an analogous experiment performed with the N-DGP.
 We thus expect the N-MOGP and the N-DGP to retain the (relative) robustness against overfitting that is characteristic of Gaussian process models.

\subsection{Missing Outputs}
\label{sec:missing}

\begin{figure}[t!]
    \centering
    \begin{subfigure}{.49\textwidth}
        \includegraphics[width=.99\textwidth]{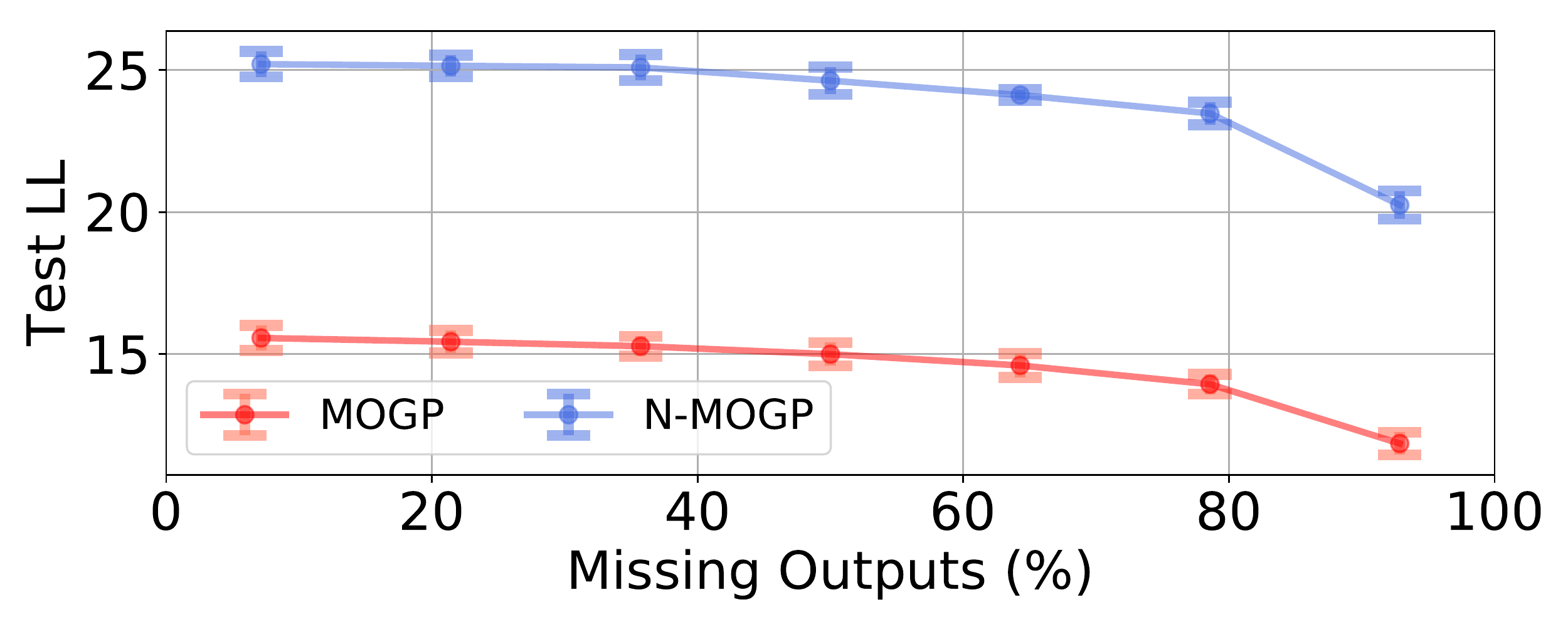}
    \end{subfigure}\hfill
        \begin{subfigure}{.49\textwidth}
        \includegraphics[width=.99\textwidth]{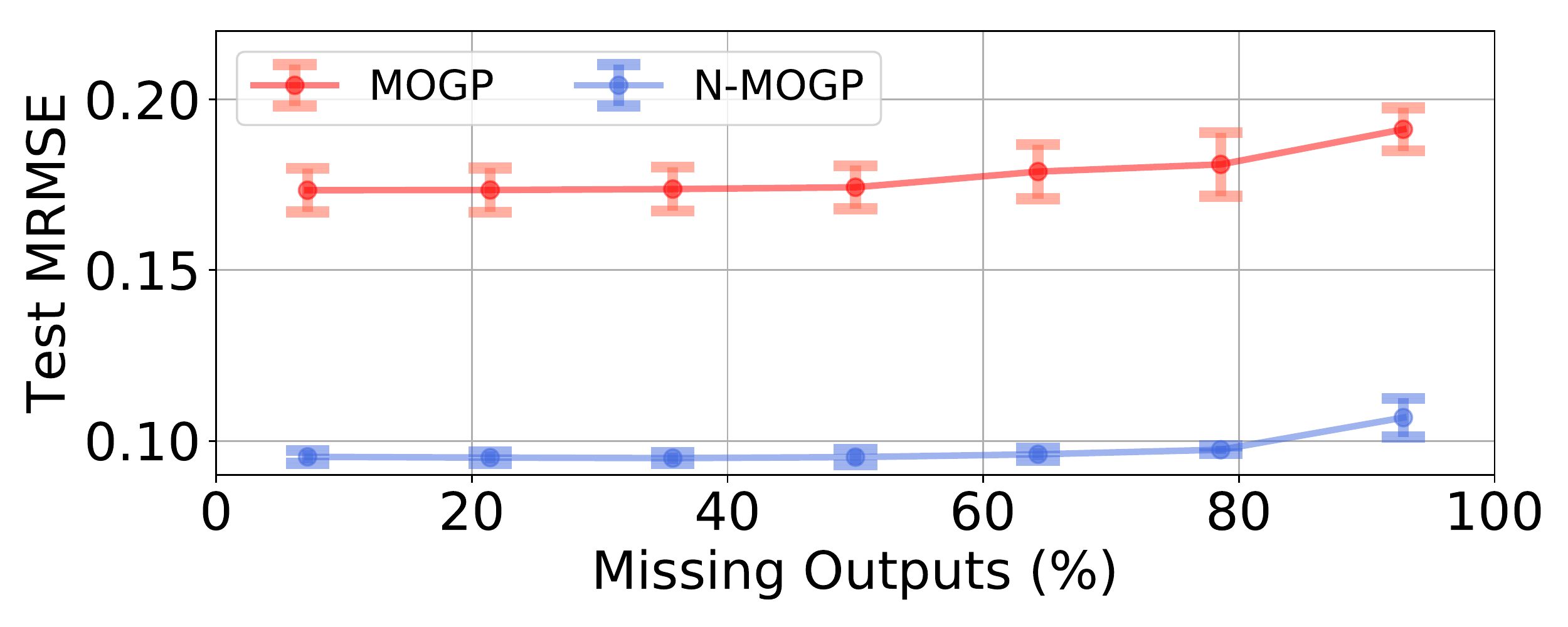}
    \end{subfigure}\hfill
        \caption{Test LLs (top) and MRMSEs (bottom) for the MOGP and N-MOGP trained with varying amounts of missing outputs $\{\by_i\}$ for the Kuka dataset. Results are averaged over ten random train/test splits. See Sec.~\ref{sec:missing} for details.}
    \label{fig:missing}
\end{figure}

Here we explore the extent to which the models defined in Sec.~\ref{sec:regmodels} can handle missing data. 
In particular we consider the case of missing outputs (i.e.~each output $\by_i$ has some number of output dimensions missing).
We compare the N-MOGP to the MOGP and report test log likelihoods and MRMSEs obtained with the Kuka dataset as we vary the number of missing output dimensions, see Fig.~\ref{fig:missing}. We find that, as is characteristic of Gaussian process models, both models maintain good performance in the presence of missing outputs.  Moreover, the N-MOGP maintains its considerable performance advantage over the MOGP over the entire percentage range of missing outputs. See the supplementary materials for similar results obtained with the F-Baxter dataset.

\section{Discussion}
\label{sec:discussion}

Neural likelihoods offer a simple and effective way to augment multi-output GP models and make them more flexible.
We expect this class of likelihoods to be most useful in scenarios where the output dimension $D_Y$ is large.
In these cases it may be impractical to consider models constructed with $L=D_Y$ Gaussian processes
so that it becomes necessary to choose $L \ll D_Y$.
In order to form a likelihood, we then need to transform the $L$-dimensional latent vector of function values $\bF(\bx)$ into $D_Y$ dimensions.
While this can be done with a simple linear transformation,
as is done in the MOGP in Sec.~\ref{sec:mogp}, it is natural to consider more flexible alternatives as represented by the
N-MOGP and SBGPRN.
Empirically, we have seen that this added flexibility can result in substantial gains in predictive performance.
Importantly, this added flexibility comes at little additional computational cost, as any computations done in the likelihood
tend to be negligible when compared to the costs associated with the Gaussian process prior. Moreover, neural likelihoods
are complementary to other methods for making GP priors flexible, as we demonstrated empirically in Sec.~\ref{sec:exp} by
combining our approach with both deep GPs and deep kernels.

There are several interesting avenues for future research. In our experiments we have focused on
regression and unsupervised learning. However, it could be of particular interest to apply neural likelihoods to the multi-task setting---for example to tasks that do not share a common set of inputs---where the additional flexibility offered by a neural likelihood could be especially beneficial. Finally, for the deterministic neural network
used to define the SBGPRN in Sec.~\ref{sec:sbgprn}, we have relied on weight decay for regularization.
It could be fruitful to explore variants of the SBGPRN that employ other techniques for regularizing neural networks, including for
example dropout \citep{srivastava2014dropout}.

\subsubsection*{Acknowledgements}

We cordially thank Ching-An Cheng for providing some of the datasets we used in our experiments.
MJ would like to thank Felipe Petroski Such for help with infrastructure for efficient distribution of experiments.


\bibliographystyle{abbrv}
\bibliography{biblio}


\clearpage

\section{Appendix}

\subsection{Additional Experimental Results}

\subsubsection{Regression}

See Table~\ref{table:mogp} for additional results for the (N-)MOGP models. Here as elsewhere the prefix `DK' indicates that the
given model is equipped with a deep kernel.
\begin{table*}[t]
\centering
\tabcolsep=0.11cm
\resizebox{1 \textwidth}{!}{
\begin{tabular}{|*{11}{c|}}
\hline
 \cellcolor[gray]{0.65}& \multicolumn{10}{c|}{\cellcolor[gray]{0.80}{\bf Dataset}} \\
   \cellcolor[gray]{0.65}& \multicolumn{2}{c|}{{\bf R-Baxter}\cellcolor[gray]{0.95}} & \multicolumn{2}{c|}{{\bf F-Baxter}\cellcolor[gray]{0.95}} & \multicolumn{2}{c|}{{\bf Kuka}\cellcolor[gray]{0.95}} & \multicolumn{2}{c|}{{\bf Sarcos}\cellcolor[gray]{0.95}} & \multicolumn{2}{c|}{{\bf MuJoCo}\cellcolor[gray]{0.95}}\\
  \Xcline{1-11}{0.5pt}
 \small {\bf Model } \cellcolor[gray]{0.80} & \tiny LL & \tiny MRMSE & \tiny LL & \tiny MRMSE & \tiny LL & \tiny MRMSE & \tiny LL & \tiny MRMSE & \tiny LL & \tiny MRMSE\\ 
\Xcline{1-11}{0.5pt}\hline\hline
\lgc  \tiny MOGP &  \llArbax & \mrmseArbax & \llAfbax & \mrmseAfbax & \llAkuka & \mrmseAkuka & \llAsarc & \mrmseAsarc & \llAmujo & \mrmseAmujo \\ \hline
\lgc  \tiny DK-MOGP &  \llArbaxdk & \mrmseArbaxdk & \llAfbaxdk & \mrmseAfbaxdk & \llAkukadk & \mrmseAkukadk & \llAsarcdk & \mrmseAsarcdk & \llAmujodk & \mrmseAmujodk  \\ \hline  \hline
    \lgc   \tiny N-MOGP (erf)&  \llMsArbaxerfdh & \mrmseMsArbaxerfdh & \llMsAfbaxerfdh & \mrmseMsAfbaxerfdh & \llMsAkukaerfdh & \mrmseMsAkukaerfdh & \llMsAsarcerfdh & \mrmseMsAsarcerfdh & \llMsAmujoerfdh & \mrmseMsAmujoerfdh \\ \hline
    \lgc   \tiny N-MOGP (relu)&  \llMsArbaxreludh & \mrmseMsArbaxreludh & \llMsAfbaxreludh & \mrmseMsAfbaxreludh & \llMsAkukareludh & \mrmseMsAkukareludh & \llMsAsarcreludh & \mrmseMsAsarcreludh & \llMsAmujoreludh & \mrmseMsAmujoreludh \\ \hline
    \lgc    \tiny N-MOGP (sherf)&  \llMsArbaxerftdh & \mrmseMsArbaxerftdh & \llMsAfbaxerftdh & \mrmseMsAfbaxerftdh & \llMsAkukaerftdh & \mrmseMsAkukaerftdh & \llMsAsarcerftdh & \mrmseMsAsarcerftdh & \llMsAmujoerftdh & \mrmseMsAmujoerftdh \\ \hline 
    \lgc   \tiny N-MOGP (leaky) &  \llMsArbaxleakyreludh & \mrmseMsArbaxleakyreludh & \llMsAfbaxleakyreludh & \mrmseMsAfbaxleakyreludh & \llMsAkukaleakyreludh & \mrmseMsAkukaleakyreludh & \llMsAsarcleakyreludh & \mrmseMsAsarcleakyreludh & \llMsAmujoleakyreludh & \mrmseMsAmujoleakyreludh \\ \hline
\end{tabular}
} 
    \caption{Full results for the (N-)MOGP models in Sec.~\ref{sec:regexp}. We report test log likelihoods per datapoint (LL) and mean root mean squared errors (MRMSE) averaged over ten random train/test splits of the data.}
  \label{table:mogp}
\end{table*}

See Table~\ref{table:gprn} for additional results for the GPRN family of models. Perhaps not surprisingly,
the DK-N-SBGPRN, which contain neural networks in both the kernel and the likelihood, is the best performing model
across the board.
\begin{table*}[t]
\centering
\tabcolsep=0.11cm
\resizebox{1 \textwidth}{!}{
\begin{tabular}{|*{11}{c|}}
\hline
 \cellcolor[gray]{0.65}& \multicolumn{10}{c|}{\cellcolor[gray]{0.80}{\bf Dataset}} \\
   \cellcolor[gray]{0.65}& \multicolumn{2}{c|}{{\bf R-Baxter}\cellcolor[gray]{0.95}} & \multicolumn{2}{c|}{{\bf F-Baxter}\cellcolor[gray]{0.95}} & \multicolumn{2}{c|}{{\bf Kuka}\cellcolor[gray]{0.95}} & \multicolumn{2}{c|}{{\bf Sarcos}\cellcolor[gray]{0.95}} & \multicolumn{2}{c|}{{\bf MuJoCo}\cellcolor[gray]{0.95}}\\
  \Xcline{1-11}{0.5pt}
 \small {\bf Model } \cellcolor[gray]{0.80} & \tiny LL & \tiny MRMSE & \tiny LL & \tiny MRMSE & \tiny LL & \tiny MRMSE & \tiny LL & \tiny MRMSE & \tiny LL & \tiny MRMSE\\ 
\Xcline{1-11}{0.5pt}\hline\hline
 \lgc  \tiny GPRN &  \llgprnrbax & \mrmsegprnrbax & \llgprnfbax & \mrmsegprnfbax & \llgprnkuka & \mrmsegprnkuka & \llgprnsarc & \mrmsegprnsarc & \llgprnmujo & \mrmsegprnmujo \\ \hline
  \lgc  \tiny DK-GPRN &  \llgprnrbaxdk & \mrmsegprnrbaxdk & \llgprnfbaxdk & \mrmsegprnfbaxdk & \llgprnkukadk & \mrmsegprnkukadk & \llgprnsarcdk & \mrmsegprnsarcdk & \llgprnmujodk & \mrmsegprnmujodk \\ \hline \hline
\lgc  \tiny SBGPRN &  \llMxrbax & \mrmseMxrbax & \llMxfbax & \mrmseMxfbax & \llMxkuka & \mrmseMxkuka & \llMxsarc & \mrmseMxsarc & \llMxmujo & \mrmseMxmujo \\ \hline
\lgc  \tiny DK-SBGPRN &  \llMxrbaxdk & \mrmseMxrbaxdk & \llMxfbaxdk & \mrmseMxfbaxdk & \llMxkukadk & \mrmseMxkukadk & \llMxsarcdk & \mrmseMxsarcdk & \llMxmujodk & \mrmseMxmujodk \\ \hline \hline
    \lgc   \tiny N-SBGPRN (erf)&  \llMsMxrbaxerf & \mrmseMsMxrbaxerf & \llMsMxfbaxerf & \mrmseMsMxfbaxerf & \llMsMxkukaerf & \mrmseMsMxkukaerf & \llMsMxsarcerf & \mrmseMsMxsarcerf & \llMsMxmujoerf & \mrmseMsMxmujoerf \\ \hline
    \lgc   \tiny N-SBGPRN (relu)&  \llMsMxrbaxrelu & \mrmseMsMxrbaxrelu & \llMsMxfbaxrelu & \mrmseMsMxfbaxrelu & \llMsMxkukarelu & \mrmseMsMxkukarelu & \llMsMxsarcrelu & \mrmseMsMxsarcrelu & \llMsMxmujorelu & \mrmseMsMxmujorelu \\ \hline
    \lgc   \tiny N-SBGPRN (sherf) &  \llMsMxrbaxerft & \mrmseMsMxrbaxerft & \llMsMxfbaxerft & \mrmseMsMxfbaxerft & \llMsMxkukaerft & \mrmseMsMxkukaerft & \llMsMxsarcerft & \mrmseMsMxsarcerft & \llMsMxmujoerft & \mrmseMsMxmujoerft \\ \hline
    \lgc   \tiny N-SBGPRN (leaky)&  \llMsMxrbaxleakyrelu & \mrmseMsMxrbaxleakyrelu & \llMsMxfbaxleakyrelu & \mrmseMsMxfbaxleakyrelu & \llMsMxkukaleakyrelu & \mrmseMsMxkukaleakyrelu & \llMsMxsarcleakyrelu & \mrmseMsMxsarcleakyrelu & \llMsMxmujoleakyrelu & \mrmseMsMxmujoleakyrelu \\ \hline\hline
    \lgc  \tiny DK-N-SBGPRN (erf)&  \llMsMxrbaxerfdk & \mrmseMsMxrbaxerfdk & \llMsMxfbaxerfdk & \mrmseMsMxfbaxerfdk & \llMsMxkukaerfdk & \mrmseMsMxkukaerfdk & \llMsMxsarcerfdk & \mrmseMsMxsarcerfdk & \llMsMxmujoerfdk & \mrmseMsMxmujoerfdk \\ \hline
    \lgc   \tiny DK-N-SBGPRN (relu)&  \llMsMxrbaxreludk & \mrmseMsMxrbaxreludk & \llMsMxfbaxreludk & \mrmseMsMxfbaxreludk & \llMsMxkukareludk & \mrmseMsMxkukareludk & \llMsMxsarcreludk & \mrmseMsMxsarcreludk & \llMsMxmujoreludk & \mrmseMsMxmujoreludk \\ \hline
    \lgc   \tiny DK-N-SBGPRN (sherf)&  \llMsMxrbaxerftdk & \mrmseMsMxrbaxerftdk & \llMsMxfbaxerftdk & \mrmseMsMxfbaxerftdk & \llMsMxkukaerftdk & \mrmseMsMxkukaerftdk & \llMsMxsarcerftdk & \mrmseMsMxsarcerftdk & \llMsMxmujoerftdk & \mrmseMsMxmujoerftdk \\ \hline
    \lgc \tiny DK-N-SBGPRN (leaky)&  \llMsMxrbaxleakyreludk & \mrmseMsMxrbaxleakyreludk & \llMsMxfbaxleakyreludk & \mrmseMsMxfbaxleakyreludk & \llMsMxkukaleakyreludk & \mrmseMsMxkukaleakyreludk & \llMsMxsarcleakyreludk & \mrmseMsMxsarcleakyreludk & \llMsMxmujoleakyreludk & \mrmseMsMxmujoleakyreludk \\ \hline 
\end{tabular}
} 
    \caption{Full results for the GPRN, SBGPRN, and N-SBGPRN models in Sec.~\ref{sec:regexp}. We report test log likelihoods per datapoint (LL) and mean root mean squared errors (MRMSE) averaged over ten random train/test splits of the data.}
  \label{table:gprn}
\end{table*}

See Table~\ref{table:dgp} for additional results for the (N-)DGP models. The models with $L^\prime=L$ employ less
flexible prios than the models for which we report results in the main text. Note that, as mentioned in the main text,
the performance gain from adding neural likelihoods is significantly larger for these models.
\begin{table*}[t]
\centering
\tabcolsep=0.11cm
\resizebox{1 \textwidth}{!}{
\begin{tabular}{|*{11}{c|}}
\hline
 \cellcolor[gray]{0.65}& \multicolumn{10}{c|}{\cellcolor[gray]{0.80}{\bf Dataset}} \\
   \cellcolor[gray]{0.65}& \multicolumn{2}{c|}{{\bf R-Baxter}\cellcolor[gray]{0.95}} & \multicolumn{2}{c|}{{\bf F-Baxter}\cellcolor[gray]{0.95}} & \multicolumn{2}{c|}{{\bf Kuka}\cellcolor[gray]{0.95}} & \multicolumn{2}{c|}{{\bf Sarcos}\cellcolor[gray]{0.95}} & \multicolumn{2}{c|}{{\bf MuJoCo}\cellcolor[gray]{0.95}}\\
  \Xcline{1-11}{0.5pt}
 \small {\bf Model } \cellcolor[gray]{0.80} & \tiny LL & \tiny MRMSE & \tiny LL & \tiny MRMSE & \tiny LL & \tiny MRMSE & \tiny LL & \tiny MRMSE & \tiny LL & \tiny MRMSE\\ 
\Xcline{1-11}{0.5pt}\hline\hline
 \lgc  \tiny DGP &  \lldeepgprbax & \mrmsedeepgprbax & \lldeepgpfbax & \mrmsedeepgpfbax & \lldeepgpkuka & \mrmsedeepgpkuka & \lldeepgpsarc & \mrmsedeepgpsarc & \lldeepgpmujo & \mrmsedeepgpmujo \\ \hline 
    \lgc  \tiny N-DGP (erf)&  \llndgprbaxerfdh & \mrmsendgprbaxerfdh & \llndgpfbaxerfdh & \mrmsendgpfbaxerfdh & \llndgpkukaerfdh & \mrmsendgpkukaerfdh & \llndgpsarcerfdh & \mrmsendgpsarcerfdh & \llndgpmujoerfdh & \mrmsendgpmujoerfdh \\ \hline 
    \lgc  \tiny N-DGP (sherf) &  \llndgprbaxerftdh & \mrmsendgprbaxerftdh & \llndgpfbaxerftdh & \mrmsendgpfbaxerftdh & \llndgpkukaerftdh & \mrmsendgpkukaerftdh & \llndgpsarcerftdh & \mrmsendgpsarcerftdh & \llndgpmujoerftdh & \mrmsendgpmujoerftdh \\ \hline 
      \lgc  \tiny N-DGP (leaky) &  \llndgprbaxleakyreludh & \mrmsendgprbaxleakyreludh & \llndgpfbaxleakyreludh & \mrmsendgpfbaxleakyreludh & \llndgpkukaleakyreludh & \mrmsendgpkukaleakyreludh & \llndgpsarcleakyreludh & \mrmsendgpsarcleakyreludh & \llndgpmujoleakyreludh & \mrmsendgpmujoleakyreludh \\ \hline \hline
    \lgc  \tiny DGP ($L^\prime=L$) &  \lldeepgprbaxsmallL & \mrmsedeepgprbaxsmallL & \lldeepgpfbaxsmallL & \mrmsedeepgpfbaxsmallL & \lldeepgpkukasmallL & \mrmsedeepgpkukasmallL & \lldeepgpsarcsmallL & \mrmsedeepgpsarcsmallL & \lldeepgpmujosmallL & \mrmsedeepgpmujosmallL \\ \hline 
    \lgc  \tiny N-DGP ($L^\prime=L$, erf)&  \llndgprbaxerfsmallL & \mrmsendgprbaxerfsmallL & \llndgpfbaxerfsmallL & \mrmsendgpfbaxerfsmallL & \llndgpkukaerfsmallL & \mrmsendgpkukaerfsmallL & \llndgpsarcerfsmallL & \mrmsendgpsarcerfsmallL & \llndgpmujoerfsmallL & \mrmsendgpmujoerfsmallL \\ \hline
    \lgc  \tiny N-DGP ($L^\prime=L$, sherf) &  \llndgprbaxerftsmallL & \mrmsendgprbaxerftsmallL & \llndgpfbaxerftsmallL & \mrmsendgpfbaxerftsmallL & \llndgpkukaerftsmallL & \mrmsendgpkukaerftsmallL & \llndgpsarcerftsmallL & \mrmsendgpsarcerftsmallL & \llndgpmujoerftsmallL & \mrmsendgpmujoerftsmallL \\ \hline
      \lgc  \tiny N-DGP ($L^\prime=L$, leaky) &  \llndgprbaxleakyrelusmallL & \mrmsendgprbaxleakyrelusmallL & \llndgpfbaxleakyrelusmallL & \mrmsendgpfbaxleakyrelusmallL & \llndgpkukaleakyrelusmallL & \mrmsendgpkukaleakyrelusmallL & \llndgpsarcleakyrelusmallL & \mrmsendgpsarcleakyrelusmallL & \llndgpmujoleakyrelusmallL & \mrmsendgpmujoleakyrelusmallL \\ \hline
\end{tabular}
} 
    \caption{Full results for the (N-)DGP models in Sec.~\ref{sec:regexp}. We report test log likelihoods per datapoint (LL) and mean root mean squared errors (MRMSE) averaged over ten random train/test splits of the data. Unless noted otherwise $L^\prime = \ceil{\tfrac{3}{4}D_Y}$.}
  \label{table:dgp}
\end{table*}

\subsubsection{Unsupervised Learning}

See Table~\ref{table:lvmsuppl} for additional results for the unsupervised learning experiments in Sec.~\ref{sec:lvmexp}.
For the N-MOGP models we consider both $D_H=7$ and $D_H=14$ hidden units.
\begin{table*}[t!]
\centering
\tabcolsep=0.11cm
\resizebox{1 \textwidth}{!}{
\begin{tabular}{|*{11}{c|}}
\hline
 \cellcolor[gray]{0.65}& \multicolumn{10}{c|}{\cellcolor[gray]{0.80}{\bf Dataset}} \\
   \cellcolor[gray]{0.65}& \multicolumn{2}{c|}{{\bf R-Baxter}\cellcolor[gray]{0.95}} & \multicolumn{2}{c|}{{\bf F-Baxter}\cellcolor[gray]{0.95}} & \multicolumn{2}{c|}{{\bf Kuka}\cellcolor[gray]{0.95}} & \multicolumn{2}{c|}{{\bf Sarcos}\cellcolor[gray]{0.95}} & \multicolumn{2}{c|}{{\bf MuJoCo}\cellcolor[gray]{0.95}}\\
  \Xcline{1-11}{0.5pt}
 \small {\bf Model}  \cellcolor[gray]{0.80} & \tiny LL & \tiny MRMSE & \tiny LL & \tiny MRMSE & \tiny LL & \tiny MRMSE & \tiny LL & \tiny MRMSE & \tiny LL & \tiny MRMSE\\ 
\Xcline{1-11}{0.5pt}
\lgc  \tiny MOGP &  \lllvmArbax & \mrmselvmArbax & \lllvmAfbax & \mrmselvmAfbax & \lllvmAkuka & \mrmselvmAkuka & \lllvmAsarc & \mrmselvmAsarc & \lllvmAmujo & \mrmselvmAmujo \\ \hline \hline
    \lgc   \tiny N-MOGP ($D_H=7$, erf)&  \lllvmMsArbaxerf & \mrmselvmMsArbaxerf & \lllvmMsAfbaxerf & \mrmselvmMsAfbaxerf & \lllvmMsAkukaerf & \mrmselvmMsAkukaerf & \lllvmMsAsarcerf & \mrmselvmMsAsarcerf & \lllvmMsAmujoerf & \mrmselvmMsAmujoerf \\ \hline
    \lgc   \tiny N-MOGP ($D_H=7$, relu)&  \lllvmMsArbaxrelu & \mrmselvmMsArbaxrelu & \lllvmMsAfbaxrelu & \mrmselvmMsAfbaxrelu & \lllvmMsAkukarelu & \mrmselvmMsAkukarelu & \lllvmMsAsarcrelu & \mrmselvmMsAsarcrelu & \lllvmMsAmujorelu & \mrmselvmMsAmujorelu \\ \hline
    \lgc    \tiny N-MOGP ($D_H=7$, sherf)&  \lllvmMsArbaxerft & \mrmselvmMsArbaxerft & \lllvmMsAfbaxerft & \mrmselvmMsAfbaxerft & \lllvmMsAkukaerft & \mrmselvmMsAkukaerft & \lllvmMsAsarcerft & \mrmselvmMsAsarcerft & \lllvmMsAmujoerft & \mrmselvmMsAmujoerft \\ \hline
    \lgc   \tiny N-MOGP ($D_H=7$, leaky)&  \lllvmMsArbaxleakyrelu & \mrmselvmMsArbaxleakyrelu & \lllvmMsAfbaxleakyrelu & \mrmselvmMsAfbaxleakyrelu & \lllvmMsAkukaleakyrelu & \mrmselvmMsAkukaleakyrelu & \lllvmMsAsarcleakyrelu & \mrmselvmMsAsarcleakyrelu & \lllvmMsAmujoleakyrelu & \mrmselvmMsAmujoleakyrelu \\\hline \hline
    \lgc   \tiny N-MOGP ($D_H=14$, erf) &  \lllvmMsArbaxerfdh & \mrmselvmMsArbaxerfdh & \lllvmMsAfbaxerfdh & \mrmselvmMsAfbaxerfdh & \lllvmMsAkukaerfdh & \mrmselvmMsAkukaerfdh & \lllvmMsAsarcerfdh & \mrmselvmMsAsarcerfdh & \lllvmMsAmujoerfdh & \mrmselvmMsAmujoerfdh \\ \hline
    \lgc   \tiny N-MOGP ($D_H=14$, relu)&  \lllvmMsArbaxreludh & \mrmselvmMsArbaxreludh & \lllvmMsAfbaxreludh & \mrmselvmMsAfbaxreludh & \lllvmMsAkukareludh & \mrmselvmMsAkukareludh & \lllvmMsAsarcreludh & \mrmselvmMsAsarcreludh & \lllvmMsAmujoreludh & \mrmselvmMsAmujoreludh \\ \hline
    \lgc    \tiny N-MOGP ($D_H=14$, sherf)&  \lllvmMsArbaxerftdh & \mrmselvmMsArbaxerftdh & \lllvmMsAfbaxerftdh & \mrmselvmMsAfbaxerftdh & \lllvmMsAkukaerftdh & \mrmselvmMsAkukaerftdh & \lllvmMsAsarcerftdh & \mrmselvmMsAsarcerftdh & \lllvmMsAmujoerftdh & \mrmselvmMsAmujoerftdh \\ \hline
    \lgc   \tiny N-MOGP ($D_H=14$, leaky)&  \lllvmMsArbaxleakyreludh & \mrmselvmMsArbaxleakyreludh & \lllvmMsAfbaxleakyreludh & \mrmselvmMsAfbaxleakyreludh & \lllvmMsAkukaleakyreludh & \mrmselvmMsAkukaleakyreludh & \lllvmMsAsarcleakyreludh & \mrmselvmMsAsarcleakyreludh & \lllvmMsAmujoleakyreludh & \mrmselvmMsAmujoleakyreludh \\\hline\hline
    \lgc   \tiny N-SBGPRN (erf) &  \lllvmMsMxrbaxerf & \mrmselvmMsMxrbaxerf & \lllvmMsMxfbaxerf & \mrmselvmMsMxfbaxerf & \lllvmMsMxkukaerf & \mrmselvmMsMxkukaerf & \lllvmMsMxsarcerf & \mrmselvmMsMxsarcerf & \lllvmMsMxmujoerf & \mrmselvmMsMxmujoerf \\ \hline
    \lgc   \tiny N-SBGPRN (relu)&  \lllvmMsMxrbaxrelu & \mrmselvmMsMxrbaxrelu & \lllvmMsMxfbaxrelu & \mrmselvmMsMxfbaxrelu & \lllvmMsMxkukarelu & \mrmselvmMsMxkukarelu & \lllvmMsMxsarcrelu & \mrmselvmMsMxsarcrelu & \lllvmMsMxmujorelu & \mrmselvmMsMxmujorelu \\ \hline
    \lgc   \tiny N-SBGPRN (sherf) &  \lllvmMsMxrbaxerft & \mrmselvmMsMxrbaxerft & \lllvmMsMxfbaxerft & \mrmselvmMsMxfbaxerft & \lllvmMsMxkukaerft & \mrmselvmMsMxkukaerft & \lllvmMsMxsarcerft & \mrmselvmMsMxsarcerft & \lllvmMsMxmujoerft & \mrmselvmMsMxmujoerft \\ \hline
    \lgc   \tiny N-SBGPRN (leaky)&  \lllvmMsMxrbaxleakyrelu & \mrmselvmMsMxrbaxleakyrelu & \lllvmMsMxfbaxleakyrelu & \mrmselvmMsMxfbaxleakyrelu & \lllvmMsMxkukaleakyrelu & \mrmselvmMsMxkukaleakyrelu & \lllvmMsMxsarcleakyrelu & \mrmselvmMsMxsarcleakyrelu & \lllvmMsMxmujoleakyrelu & \mrmselvmMsMxmujoleakyrelu \\ \hline
\end{tabular}
} 
\caption{Full results for the unsupervised learning experiments in Sec.~\ref{sec:lvmexp}. We report test log likelihoods per datapoint (LL) and mean root mean squared errors (MRMSE) averaged over ten random train/test splits of the data.}
  \label{table:lvmsuppl}
\end{table*}

\subsection{Experimental Details}

For all experiments we use the Adam optimizer \citep{kingma2014adam}.

\subsubsection{Synthetic Experiment}

We follow the training protocol discussed in the next section.

\subsubsection{Regression}

We specify some of the details of our regression models and their corresponding inference procedures
omitted in the main text.
Our RBF kernels use separate length scales for each input dimension.
For all models we choose $N_{\rm ind}=400$ inducing points,
except for the GPRN (where we choose $N_{\rm ind}=100$) and for the \mbox{N-DGP} models
(where we choose $N_{\rm ind}=400$ for the first layer of GPs and $N_{\rm ind}=100$ for the second layer of GPs). We find that these models can struggle to take advantage of more inducing points and become susceptible to getting stuck in bad local optima when the number of inducing points is too large.
For the results in Table~\ref{table:reg} we choose the shifted erf non-linearity for the N-MOGP, the leaky ReLU non-linearity for the N-SBGPRN and DK-N-SBGPRN, and the erf non-linearity for the N-DGP.

We train all models for 250 epochs. We use mini-batch sizes of 1000, 500, 500, 500, and 250 for the
MuJoCo, Kuka, F-Baxter, Sarcos, and R-Baxter datasets, respectively, except for the (N-)DGP models, where
we double the mini-batch size. For each training/test split we use 5 random parameter initializations and only train the best performing model (in terms of training LL) to completion.
Depending on the model we use initial learning rates in the range $[0.01, 0.05]$, which we reduce stepwise over the course of training. For some of the (N-)DGP models we also found it useful to employ KL annealing during the first 20 epochs of training. Except for the inducing points for the second layer of GPs in the DGP (which we initialize randomly and where we do not share the inducing points across the $L^\prime$ GPs), we initialize inducing points using k-means clustering on the training data inputs $\{ \bx_i\}$.

For uniformity we train all models except for the GPRN\footnote{We train the GPRN by computing analytic expected log likelihoods as outlined in Sec.~\ref{sec:analyticelbo}.}
using the SGVB approach outlined in Sec.~\ref{sec:sgvb}. That is, we compute
stochastic gradient estimates of the expected log likelihood by drawing $N_{\rm samples}=250$ samples from the
variational distributions $\{q_\ell(\bff_{\rm mb}) \}$ for each GP. For the (N-)DGP models, where the nesting of multiple layers of Gaussian processes makes sampling particularly expensive, we instead use
$N_{\rm samples}=5$ samples.\footnote{Our inference procedure for the (N-)DGP models closely follows that of \citep{salimbeni2017doubly}.} Note that for models that make use of a mixing matrix $\bM$ we simply integrate $\bM$ out and never sample $\bM$.

Throughout we report test log likelihoods using an estimator of the form
\begin{equation}
\nonumber
\begin{split}
&\log p(\bY^* | \bX^*) \approx \\
&\tfrac{1}{N_{\rm outer}} \Sigma_{k=1}^{N_{\rm outer}}
\log \left( \tfrac{1}{N_{\rm inner}}\Sigma_{j=1}^{N_{\rm inner}} p(\bY^* | \bX^*, \bF_{jk}^*)\right)
\end{split}
\end{equation}
where each sample $\bF_{jk}^*$ is from the relevant variational distribution and
$N_{\rm outer}=25$ and $N_{\rm inner} = 50$.

\subsubsubsection{{\bf Bias in Hidden Units}}

For the N-MOGP, N-SBGPRN, and N-DGP we use $D_H$ bias terms in the non-linearity $\sigma$ that appears in the likelihood. That is, the expression $\sigma(\bMt \bF)$ in Eqn.~\ref{eqn:nmogp}
is in fact shorthand for $\sigma(\bMt \bF + \bm{b})$, where
$\bm{b}$ is a $D_H$-dimensional vector of bias terms with a unit Normal prior. During inference, we use mean field Normal variational distributions for each bias term $b_h$. For these same models, because of the flexibility of the bias units, we use Gaussian priors with (fixed) zero mean functions, while for all other Gaussian process priors we use (trainable) constant mean functions.

\subsubsubsection{{\bf Deep Kernels}}

All our deep kernels make use of neural networks with two hidden layers, each with 50 hidden units, and have the same number of output dimensions as input dimensions. We use tanh non-linearities. We also use a multiplicative parameterization of the deep kernels such that at initialization each is close to the identity function.

\subsubsubsection{{\bf SBGPRN Neural Networks}}

Throughout the neural network $\bM(\bx)$ that appears in the SBGPRN and N-SBGPRN has two hidden layers, each with 50 units, and utilizes tanh non-linearities.

\subsubsection{Unsupervised Learning}

We specify some of the details of our unsupervised models and their corresponding inference procedures
omitted in the main text.
We use $N_{\rm ind}=200$ inducing points for all models.
For the results in Table~\ref{table:lvm} we choose the leaky ReLU non-linearity for the N-MOGP and the ReLU non-linearity for the N-SBGPRN. The training procedure and modeling setup generally follows that of the regression models, with the important difference that (except for the latent inputs, which we always sample) we compute analytic expected log likelihoods as described in Sec.~\ref{sec:analyticelbo}.
We use $N_{\rm qp}=100$ quadrature points.
During training we sample a single latent $\bx_i \sim q(\bx_i)$ for each datapoint.
We find that using analytic ELBOs during training leads to better stability and performance. In contrast to the regression models, the RBF kernels in our unsupervised learning experiments use the same length scale for all dimensions. During test time, we introduce a new variational distribution $q(\bX^*)$ for the unseen data and fit
$q(\bX^*)$ by maximizing the ELBO. That is, fitting $q(\bX^*)$ proceeds analogously to training, except that now everything except for $q(\bX^*)$ is kept fixed (i.e.~the kernel hyperparameters, the variational distributions
$q(\bu_\ell)$, etc.). We initialize $q(\bX^*)$ by using a nearest neighbor algorithm to find points in the training data that are close to points in the test data; we then initialize the means of $q(\bx_i^*)$ by using the mean of the variational distribution $q(\bx_j)$ for the training datapoint $\bx_j$ that is closest to $\bx_i^*$.

\subsubsection{Varying $D_H$}

The experimental protocol for these experiments closely follows that of the regression experiments in Sec.~\ref{sec:regexp}.

\subsubsection{Small Data Regime}

The experimental protocol for these experiments closely follows that of the regression experiments in Sec.~\ref{sec:regexp},
with the difference that we only use $N_{\rm ind} =250 $ inducing points and that as we reduce the training set size we reduce
the mini-batch size proportionally.

\subsubsection{Missing Outputs}

\begin{figure}[t!]
    \centering
    \begin{subfigure}{.49\textwidth}
        \includegraphics[width=.99\textwidth]{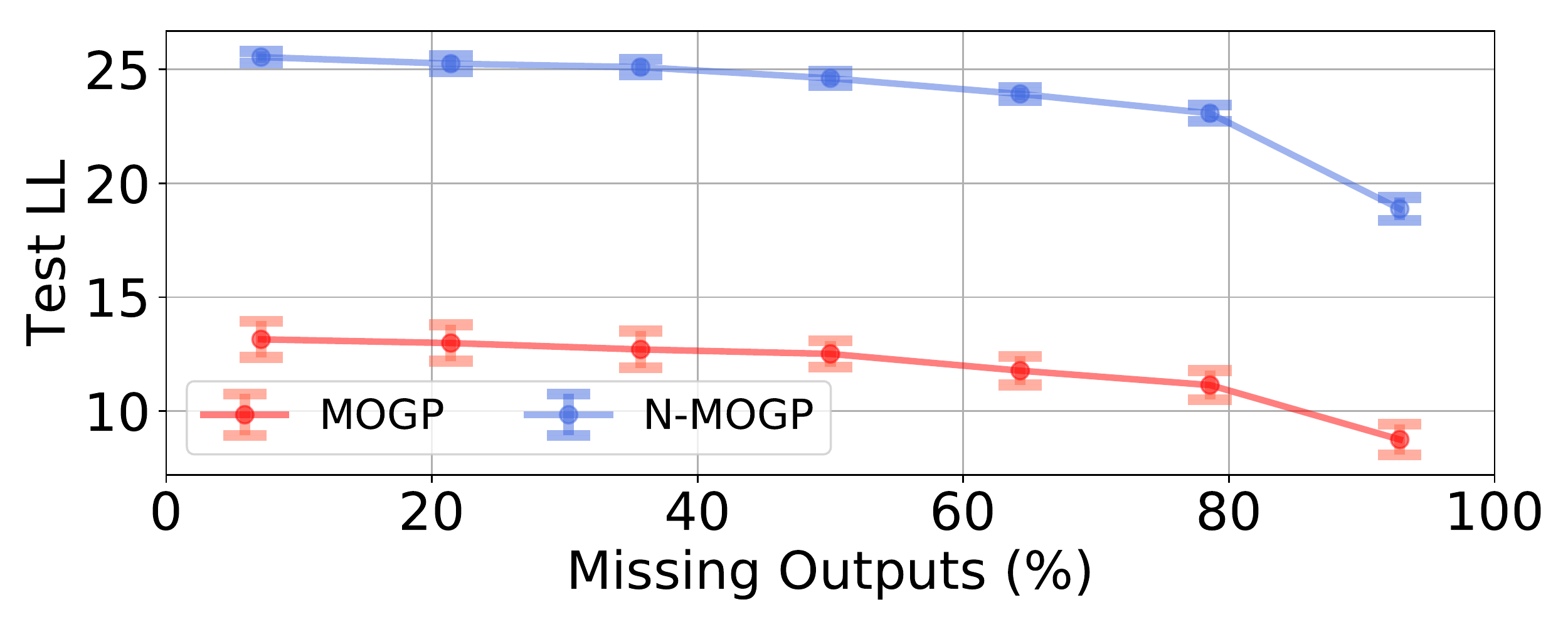}
    \end{subfigure}\hfill
        \begin{subfigure}{.49\textwidth}
        \includegraphics[width=.99\textwidth]{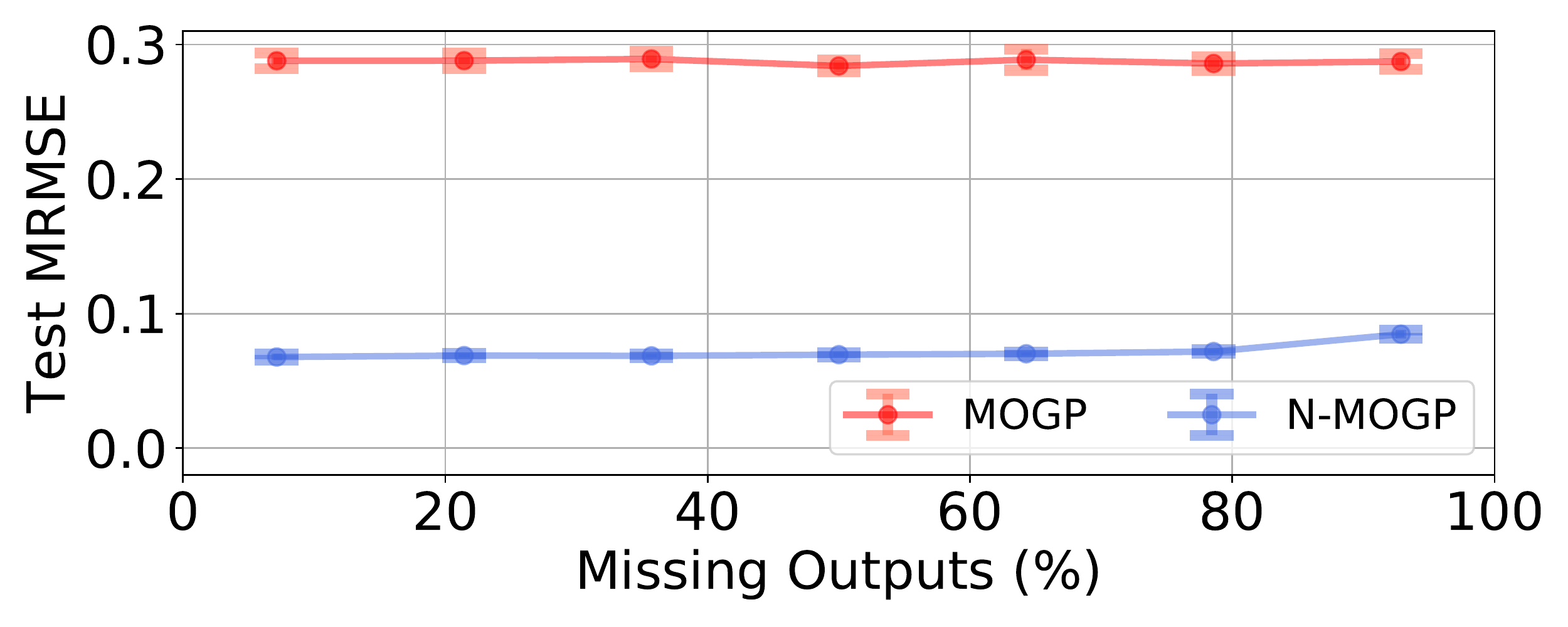}
    \end{subfigure}\hfill
        \caption{Test LLs (top) and MRMSEs (bottom) for the MOGP and N-MOGP trained with varying amounts of missing outputs $\{\by_i\}$ for the F-Baxter dataset. Results are averaged over ten random train/test splits.}
    \label{fig:missingbaxter}
\end{figure}

The experimental protocol for these experiments closely follows that of the regression experiments in Sec.~\ref{sec:regexp}.
For each missing output percentage the number of missing output dimensions for each output $\by_i$ is identical (e.g.~3/14 output dimensions are missing).
The particular missing output dimensions for each datapoint are different between each train/test split. For additional results obtained with the F-Baxter dataset see Fig.~\ref{fig:missingbaxter}.

\subsection{Expectations of Non-linearities}

We discuss how we compute the expectations required to form analytic expected log likelihoods as described for the N-MOGP in Sec.~\ref{sec:analyticelbo}. Here we focus on the error function non-linearity $g(x)=\erf(x)$. An analytic expression for the mean function can be obtained from a table of integrals:\footnote{See e.g.~the integrals listed in \citep{ng1969table}}
\begin{equation}
\label{eqn:erfmean}
\mathbb{E}_{\NN(x|\mu, \sigma)} \left[ \erf(x)  \right] =
\erf\left(\frac{ \mu}{\sqrt{1+2 \sigma^2}}\right)
\end{equation}
An analytic expression for the corresponding second moment is not readily available but can be computed efficiently using quadrature:
\begin{equation}
\begin{split}
&\mathbb{E}_{\NN(x|\mu, \sigma)} \left[ \erf(x)^2 \right] = \mathbb{E}_{\NN(x|0,1/{\sqrt{2}})} \left[ \erf(\sqrt{2} \sigma x + \mu)^2 \right] \\
 &= \frac{1}{\sqrt{\pi}}\int dx e^{-x^2} \erf(\sqrt{2} \sigma x + \mu)^2  \\
 & \approx  \frac{1}{\sqrt{\pi}}\sum_{i=1}^{N_q} w_i \erf(\sqrt{2} \sigma x_i + \mu)^2
 \end{split}
\end{equation}
where  $\{ (x_i, w_i) \}$ are the sample points and weights from a Gauss-Hermite quadrature rule of order $N_q$ (conventionally defined w.r.t.~the weighting function $e^{-x^2}$).  Finally, we would like to compute the bivariate expectation
\begin{equation}
\label{eqn:bivariate}
\mathbb{E}_{\NN(\bx |\bmu, \bSig)} \left[ g(x_1)g(x_2)  \right]  = \mathbb{E}_{\NN(\bx |\bmu, \bSig)} \left[ \erf(x_1)\erf(x_2)  \right]
\end{equation}
An analytic expression is not readily available but we can do ``half'' of the integral analytically and compute the remaining univariate integral using quadrature. Changing variables so that the Normal distribution has a diagonal covariance matrix, we obtain:
\begin{equation}
\frac{1}{\sqrt{\pi}}\! \int dx_1 e^{-x_1^2} g(\sqrt{2} L_{11} x_1 + \mu_1) h(x_1)
\end{equation}
where $\bm{L}$ is the Cholesky decomposition\footnote{Since $\bSig$ is two-dimensional this decomposition is trivial to compute: $L_{11} = \sqrt{\Sigma_{11}}$, etc.} of $\bSig$ with $\bSig=\bm{L}\bm{L}^{\rm T}$ and $h(x_1)$ is given by the mean function
\begin{equation}
\begin{split}
h(x_1)&=\mathbb{E}_{\NN(x_2|0,1/{\sqrt{2}})} \left[ g(\sqrt{2} L_{21} x_1+ \sqrt{2} L_{22} x_2 + \mu_2) \right]  \\
&=\mathbb{E}_{\NN(\tilde{x}_2|\sqrt{2}L_{21} x_1 + \mu_2,L_{22})} \left[ g(\tilde{x}_2) \right]
 \end{split}
\end{equation}
Consequently whenever an analytic expression is available for this inner expectation---as is the case for the error function, recall~Eqn.~\ref{eqn:erfmean}---the bivariate expectation in Eqn.~\ref{eqn:bivariate} can be efficiently computed with univariate Gauss-Hermite quadrature.

The identity Eqn.~\ref{eqn:erfmean} can be manipulated to yield all expectations of the form
$\mathbb{E}_{\NN(x|\mu, \sigma)} \left[ x^n {\rm erf}(x) \right]$, thus making all the mean functions for all nonlinearities of the form
$g(x) = {\rm poly}(x) \erf(x)$ for some polynomial ${\rm poly}(x)$ analytically tractable. For example we have
\begin{equation}
\begin{split}
\mathbb{E}_{\NN(x|\mu, \sigma)} \left[ x\erf(x)  \right] =
&\mu \erf\left(\frac{ \mu}{\sqrt{1+2 \sigma^2}}\right)  + \\
& \frac{2\sigma^2}{\sqrt{\pi}(1+2\sigma^2)^{1/2}} e^{-\frac{\mu^2}{1 +2\sigma^2}}
\end{split}
\end{equation}
and
\begin{equation}
\begin{split}
\mathbb{E}_{\NN(x|\mu, \sigma)} \left[ x^2\erf(x)  \right] =
&(\mu^2 + \sigma^2) \erf\left(\frac{ \mu}{\sqrt{1+2 \sigma^2}}\right)  + \\
& \frac{4\mu\sigma^2(1+\sigma^2)}{\sqrt{\pi}(1+2\sigma^2)^{3/2}} e^{-\frac{\mu^2}{1 +2\sigma^2}}
\end{split}
\end{equation}
For the analytic expressions needed to deal with piecewise polynomial non-linearities
like ReLU we refer the reader to \citep{jankowiak2018closed}.

Finally, note that above (e.g.~in Eqn.~\ref{eqn:erfmean}) we have given analytic expressions for
expectations with respect to 1-dimensional Normal random variables. However,
in Sec.~\ref{sec:nmogpinf}  the mean and variance functions for the activations,
$\bmean^\sigma(\bx)$  and $\bvar^\sigma(\bx)$, are expressed as expectations with respect to the
product distribution $\prod_\ell q(f_{\ell,i})$.  Since, however, the argument of the non-linearity $\sigma(\cdot)$
in these equations (see e.g.~Eqn.~\ref{eqn:nmogpmean}) is a linear combination of 1-dimensional Normal random variables, the
argument is in fact a 1-dimensional Normal random variable, and so our analytic results are directly applicable. We
simply appeal to the fact that if $a_i \sim \mathcal{N}(\mu_i, \sigma_i)$ and $x\equiv \sum_i b_i a_i$ for some
constants $\{ b_i \}$, then $x\sim \mathcal{N}(\mu, \sigma)$ with $\mu = \sum_i b_i \mu_i$ and
$\sigma^2 = \sum_i b_i^2 \sigma_i^2$.

\end{document}